\documentclass[10pt,twocolumn,twoside]{IEEEtran}
\usepackage{amssymb}
\usepackage{amsmath}
\usepackage{amsfonts}
\usepackage{epsfig}
\usepackage{graphicx}
\usepackage{subfigure}
\usepackage{algorithmic}
\usepackage{algorithm}
\usepackage{cite}
\usepackage{multirow}
\usepackage{times}
\usepackage{xcolor}
\usepackage{soul}
\usepackage{mathrsfs}
\usepackage{amssymb}
\usepackage{graphicx}
\usepackage{psfrag}
\usepackage{subfigure}
\usepackage{stfloats}
\usepackage{float}
\usepackage{array}
\usepackage{amsthm}
\usepackage{booktabs}  
\usepackage{algorithm} 
\usepackage{algorithmic} 
\usepackage{amsmath,amsfonts,amssymb,amsbsy,bm,paralist,theorem,ifthen,color}

\begin{document}

\title{A Self-paced Regularization Framework for Partial-Label Learning}

\author{Gengyu Lyu, Songhe Feng, Congyan Lang
\thanks{Gengyu Lyu, Songhe Feng and Congyan Lang are with School of Computer and Information Technology, Beijing Jiaotong University, Beijing, 100044, China (e-mail: \{lvgengyu, shfeng, cylang\}@bjtu.edu.cn).}
}

\markboth{SUBMITTED TO IEEE TRANSACTIONS ON Cybernetics, 2018}%
{Lyu \MakeLowercase{\textit{et al.}}: A Self-paced Regularization Framework for Partial-Label Learning}
\maketitle

\begin{abstract}
Partial label learning (\emph{PLL}) aims to solve the problem where each training instance is associated with a set of candidate labels, one of which is the correct label. Most \emph{PLL} algorithms try to disambiguate the candidate label set, by either simply treating each candidate label equally or iteratively identifying the true label. Nonetheless, existing algorithms usually treat all labels and instances equally, and the complexities of both labels and instances are not taken into consideration during the learning stage. Inspired by the successful application of self-paced learning strategy in machine learning field, we integrate the self-paced regime into the partial label learning framework and propose a novel \textbf{S}elf-\textbf{P}aced \textbf{P}artial-\textbf{L}abel \textbf{L}earning (\textbf{SP-PLL}) algorithm, which could control the learning process to alleviate the problem by ranking the priorities of the training examples together with their candidate labels during each learning iteration. Extensive experiments and comparisons with other baseline methods demonstrate the effectiveness and robustness of the proposed method.
\end{abstract}


\section{Introduction}
As a weakly-supervised learning framework, partial-label learning \footnote{In some literature, partial-label learning is also named as \emph{superset label learning} \cite{Liu:lotsllp-ICML2014},  \emph{ambiguous label learning}\cite{Chen:allud-IEEET2014} or \emph{soft label learning} \cite{Oukhellou2009Learning}.} (PLL) learns from ambiguous labeling information where each training example is associated with a candidate label set instead of a uniquely explicit label \cite{zhang:dfpll-IEEET2017}\cite{Cour:lfpl-JMLR2011} \cite{zhou2017brief}. It aims at disambiguating the ground-truth label from the candidate label set among which the other labels are incorrect.

\begin{figure}
\centering
\setlength{\abovecaptionskip}{0.cm}
\setlength{\belowcaptionskip}{0.cm}
\includegraphics[width = 3in,height=1.5in]{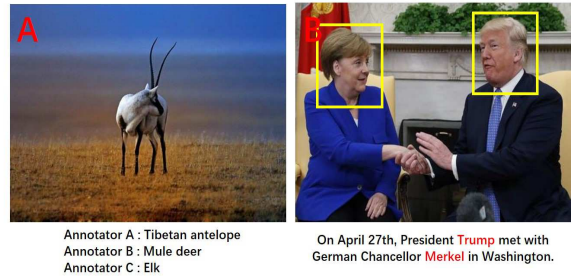}
\vspace{1mm}
\caption{Examplar applications of partial-label learning.}
\label{fig0}
\vspace{-5mm}
\end{figure}

In recent years, such learning mechanism has been widely used in many real-world scenarios. For example, in crowdsourcing online annotation system(Figure \ref{fig0}(A)) \cite{Luo:lfcls-NIPS2010}, users with different knowledge background probably annotate the same image with different labels. Thus, it is necessary to find the correspondence between each image and its ground-truth label which is resided in the candidate annotations. Another representative application is naming faces in images using text captions (Figure \ref{fig0}(B)) \cite{Chen:lfalfi-TPAMI2017} \cite{xiao2015automatic}. In this setting, since the name of the depicted people typically appears in the caption, the resulting set of images is ambiguously labeled if more than one name appears in the caption. In other words, the specific correspondence between the faces and their names are unknown. Partial label learning provides an effective solution to resolve such weakly supervision problem by disambiguating the correct label from large number of ambiguous labels. In addition to the applications mentioned above, PLL has also been widely used in many other scenarios, including web mining \cite{Luo:lfcls-NIPS2010}, facial age estimation \cite{Zhang:pllvfad-TKDD2016}, multimedia content analysis \cite{Zeng:lbaali-CVPR2013} \cite{xie2018pmll}, ecoinformatics \cite{Liu:acmmmfsll-NIPS2012}, etc.

\subsection{Related Work}
Existing PLL algorithms can be roughly grouped into the following three categories: \emph{Average Disambiguation Strategy}, \emph{Identification Disambiguation Strategy} and \emph{Disambiguation-Free Strategy}, respectively.
\subsubsection{Average Disambiguation Strategy (ADS)}
\label{section1.1}
ADS based PLL methods assume that each candidate label contributes equally to the modeling process and make prediction for unseen instances by averaging the output from all candidate labels. Following such strategy, \cite{Huller:lfale-LNCS2005} and \cite{tang:crdpll-AAAI2017} adopt an instance-based model following $\arg\max_{y\in\mathcal{Y}}\sum_{i\in\mathcal{N}_{({x}^{*})}}\!\!\mathbb{I}(y\!\!\in\!\!S_i)$ to predict the label $y^{*}$ of unseen instance $\textbf{x}^{*}$. \cite{Cour:lfpl-JMLR2011} disambiguates the ground-truth label by averaging the outputs from all candidate labels, i.e. $\frac{1}{S_i}\!\sum_{y\in S_i}F(\textbf{x},{\bm{\Theta}},y)$. \cite{Zhang:stpllpaiba-IJCAI2015} and \cite{Chen2017A} also adopt an instance-based model and they make prediction via \emph{k}-nearest neighbors weighted voting and minimum error reconstruction criterion. \cite{Zhang:pllvfad-TKDD2016} proposes the so-called \emph{PL-LEAF} algorithm, which facilitates the disambiguation process by taking the local topological information from feature space into consideration. Obviously, the ADS based PLL methods mentioned above are intuitive and easy to implement. However, these algorithms share a critical shortcoming that the output of the ground-truth label could be overwhelmed by the outputs of the other false positive labels, which will enforce negative influence on the effectiveness of the final model.

\subsubsection{Identification Disambiguation Strategy (IDS)}
IDS based PLL methods are proposed to alleviate the shortcoming of ADS based methods mentioned in Section \ref{section1.1}. This strategy aims at directly identifying the ground-truth label from corresponding candidate label set instead of averaging the output from all candidate labels. Existing PLL algorithms following this strategy often regard the ground-truth label as a latent variable first, identified as $\arg\max_{y\in{S_i}}F(\textbf{x},{\bm{\Theta}},y)$, and then refine the model parameter $\bm{\Theta}$ iteratively by utilizing some specific criterions. For example, considering the fact that treating each candidate label equally is inappropriate, Jin et al. \cite{Jin:lwml-NIPS2003} utilizes Expectation-Maximization (EM) procedure to optimize the latent variable $y$, which is based on maximum likelihood criterion: $\sum_{i=1}^{n}\log (\sum_{y\in{S_i}}F(\textbf{x},{\bm{\Theta}},y))$. Similarly, \cite{Chen:allud-IEEET2014}, \cite{Liu:acmmmfsll-NIPS2012}, \cite{Grandvalet:lfplwme-CWP2004}, \cite{Zhou2016Partial}, \cite{Jin:lwml-NIPS2003} and \cite{vannoorenberghe2005partially} also adopt the maximum likelihood criterion to refine the latent variable. Moreover, Maximum margin technique is also widely employed as objective function in the PLL problem. For example, \cite{Nguyen:cwpl-KDDM2008} maximums the margin between the output from candidate labels and non-candidate labels to train a multi-class classifier: $\sum_{i=1}^{n}(\max_{y\in{S_i}}F(\textbf{x},{\bm{\Theta}},y)-\max_{y\not\in{S_i}}F(\textbf{x},{\bm{\Theta}},y))$, while \cite{Yu:mmpll-ML2015} directly maximums the margin between the ground-truth label and the other candidate labels: $\sum_{i=1}^{n}(F(\textbf{x},{\bm{\Theta}},\emph{y})-\max_{y\neq\emph{y}}F(\textbf{x},{\bm{\Theta}},y))$. Although these IDS based PLL methods have achieved satisfactory performances in many scenarios, they suffer from the common shortcoming that training instances may be assigned with incorrect labels during each iterative optimization, especially for the PL data whose instances or candidate labels are difficult to disambiguate, and it will affect the optimization of classifier parameters in the next iteration.

\begin{figure}
\centering
\setlength{\abovecaptionskip}{0.cm}
\setlength{\belowcaptionskip}{0.cm}
\includegraphics[width = 2.75in,height=2.4in]{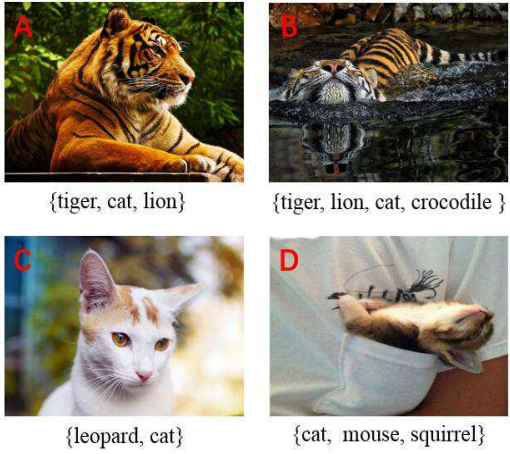}
\vspace{1mm}
\caption{Different complexity of partial-label instances}
\label{fig:one}
\vspace{-5mm}
\end{figure}

\subsubsection{Disambiguation-Free Strategy (DFS)}
More recently, different from the above two PLL strategies, some attempts have been made to learn from PL data by fitting the PL data to existing learning techniques instead of disambiguation. \cite{zhang:dfpll-IEEET2017} proposes a disambiguation-free algorithm named \emph{PL-ECOC}, which utilizes \emph{Error-Correcting Output Codes} (ECOC) coding matrix \cite{dietterich1994solving} and transfers the PLL problem into binary learning problem. \cite{TEBDfPLL-IJCAI2018} proposes another disambiguation-free algorithm called \emph{PALOC}, which enables binary decomposition for PL data in a more concise manner without relying on extra manipulations such as coding matrix. However, the performance of these two algorithms are inferior to IDS based methods in some scenarios.

\subsection{Our Motivation}
Although the algorithms mentioned above have obtained desirable performance in many real-world scenarios, they still suffer from several common drawbacks. For example, the PLL methods mentioned above treat all the training examples equally together with their candidate labels, but none of them take the complexity of training examples and that of the labels into consideration. However, in real-world scenarios, examples with different backgrounds and labels in different candidate label sets often express varying difficulties. For example, as shown in Figure \ref{fig:one}, we can easily see that images (\emph{B,D}) is harder than images (\emph{A,C}) not only from the complexity of instance but also from the number of candidate labels. In particular, when we utilize the iterative optimization method to refine the model parameters, the label \emph{cat} may be assigned to the image \emph{B} in an iteration, but obviously such assignment is a large noise in the subsequent iterations, which has bad effect on the classifier model. Thus, to improve the effectiveness of model, the complexity of the training instances together with the candidate labels should be taken into consideration.

In recent years, inspired by the cognitive process of human, \emph{Self-Paced Learning (SPL)} is proposed to deal with the above problem, which automatically leads the learning process from easy to hard \cite{Kumar:splflvm-NIPS2010}\cite{Meng:wodsplio-CS2015}. Concretely, during the optimization process of SPL, the 'easy' samples will be selected and learned first in the previous iterations, and then the 'hard' samples can be gradually selected in the subsequent iterations. The learning mechanism can smoothly guide the learning process to pay more attention to the reliable discriminative data rather than the confusing ones \cite{Pi:spblfc-IJCAI2016}. So far, SPL has obtained empirical success in many research fields, such as multi-instance learning \cite{Zhang:csdvspmilf-TPAMI2017}\cite{Sang:spdlfwsod-CVPR2017}, multi-label learning \cite{Li:asprffml-NNLS2016}, multi-task learning \cite{Li:spmtl-AAAI2017}, multi-view learning \cite{Xu:mvsplfc-ICAI2015}, matrix-factorization \cite{Zhao:splfmf-AAAI2015}, face identification \cite{Lin:asplfceapfi-TPAMI2017} and so on.

In light of this observation, in this paper, we build a connection between the PLL and the SPL, and propose a novel unified framework \emph{\textbf{S}elf-\textbf{P}aced \textbf{P}artial-\textbf{L}abel \textbf{L}earning} (\textbf{SP-PLL}). With an adaptive step from 'easy' to 'hard', SP-PLL can dynamically adjust the learning order of the training data (i.e. examples together with their candidate labels) and guide the learning to focus more on the data with high-confidence. Benefiting from the self-controlled sample-selected scheme, SP-PLL can effectively capture the valuable label information from the true label and minimize the negative impact of other candidate labels. Experimental results on UCI data sets and real-world data sets demonstrate the effectiveness of the proposed approach.

\section{Background}
In the following two sections, we separately give brief introduction about the work of partial-label learning \cite{Yu:mmpll-ML2015} and self-paced learning \cite{Zhao:splfmf-AAAI2015}, which our approach originates from.
\subsection{Partial-label learning (PLL)}
\label{Sec:background_partial_label}
Formally speaking, we denote by $\mathcal{X}\!=\!\mathbb{R}^{d}$ the \emph{d}-dimensional input space, and $\mathcal{Y}\!=\! \{1,2,\ldots,\emph{q}\}$ the output space with q class labels. PLL aims to learn a classifier $f:\mathcal{X}\mapsto\mathcal{Y}$ from the PL training data $\mathcal{D}=\{(\textbf{x}_i,S_i)\} (1\leq{i}\leq{n})$, where the instance $\textbf{x}_{i}\in\mathcal{X}$ is described as a \emph{d}-dimensional feature vector and the candidate label set $S_{i}\subseteq\mathcal{Y}$ is associated with the instance $\textbf{x}_{i}$. Furthermore, let $\textbf{y}=\{y_1,y_2,\ldots,y_n\}$ be the ground-truth label assigments for training instances and each $y_{i}\in S_{i}$ of $\textbf{x}_{i}$ is not directly accessible during the training phase.

In our algorithm, we adopt the \emph{\textbf{M}axi\textbf{M}um-\textbf{M}argin \textbf{P}artial-\textbf{L}abel learning} (\emph{M3PL}) \cite{Yu:mmpll-ML2015} to design the SP-PLL framework. Given the parametric model ${\bm{\Theta}}=\{(\textbf{w}_{p},b_{p})|1\leq p\leq q\}$ and the modeling output $F(\textbf{x}_i,{\bm{\Theta}},y)$ of $\textbf{x}_{i}$ on label $y$, different from other maximum-margin PLL algorithms, M3PL focuses on differentiating the output from ground-truth label against the maximum output from all other labels (i.e.$F(\textbf{x}_{i},{\bm{\Theta}},y_{i})-\max_{\tilde{y_i}\not= y_i}F(\textbf{x}_i,{\bm{\Theta}},\tilde{y_i})$), instead of the maximum output from candidate labels against that from non-candidate labels (i.e. $\max_{y_i\in{S_i}}F(\textbf{x}_{i},{\bm{\Theta}},y_{i})-\max_{\tilde{y_i}\not\in{S_i}}F(\textbf{x}_i,{\bm{\Theta}},\tilde{y_i})$), which can avoid the negative effect produced by the noisy labels in candidate label set. M3PL deals with the task of PLL by solving the following optimization problem \textbf{(OP1)}:
{\setlength\abovedisplayskip{10pt}
\setlength\belowdisplayskip{10pt}
\begin{flalign*}
&\min\limits_{\bm{\Theta},{\xi},\textbf{y}}\quad \frac{1}{2}\sum\limits_{p=1}^{q}\|\textbf{w}_{p}\|^{2}+\mathcal{C}\sum\limits_{i=1}^{n}\xi_{i}\\
&s.t. :\quad
\begin{cases} (\textbf{w}_{y_i}^{\top}\cdot\textbf{x}_i+b_{y_i})-\max\limits_{\tilde{y_i}\not= y_i}(\textbf{w}_{\tilde{y_i}}^{\top}\cdot\textbf{x}_i+b_{\tilde{y_i}})\geq{1-\xi_i}\\
{\xi}_{i}{\geq}{0},  \forall{i\in\{1,2,\ldots,n\}}\\
\textbf{y}\in\mathcal{S}\\
\sum_{i=1}^{n}\mathbb{I}(y_i=p)=n_p,  \forall{p\in\{1,2,\ldots,q\}}\\
\end{cases}
\end{flalign*}}where $C$ is the regularization parameter, $\bm{\xi}=\{\xi_1,\xi_2,\ldots,\xi_n\}$ is the slack variables set, $n_p$ is the prior number of examples for the \emph{p}-th class label in $\mathcal{Y}$, and $\mathcal{S}$ is the feasible solution space. $\mathbb{I}(\bigtriangleup)$ is an indicator function where $\mathbb{I}(\bigtriangleup)=1$ if and only if $\bigtriangleup$ is true, otherwise $\mathbb{I}(\bigtriangleup)=0$.

Note that \textbf{(OP1)} is a mixed-typed variables optimization problem, which needs to optimize the integer variables $\textbf{y}$ and the real-valued variables $\bm{\Theta}$ simultaneously, and \textbf{(OP1)} could be solved by using an alternating optimization procedure in an iterative manner. However, during each iterative optimization process of M3PL, the (unknown) assigned label $y_i$ is not always the true label for each instance, and the training instances assigned with such unreliable label will have negative effect on the optimization of $\bm{\Theta}$. In such case, the effectiveness and robustness of model cannot be guaranteed.

\subsection{Self-paced learning (SPL)}
In this subsection, we first define notations and then introduce the self-paced function.

We denote $\textbf{v}=\{v_1,v_2,\ldots,v_n\}$ as \emph{n}-dimensional weight vector for the $\emph{n}$ training examples, $L(\textbf{x}_{i},\textbf{w},b,y_i)$ as the empirical loss for the \emph{i}-th training example, and $\lambda$ as the self-paced parameter for controlling the learning process. The general self-paced framework could be designed as: \cite{Kumar:splflvm-NIPS2010} \cite{meng2017theoretical}
{\setlength\abovedisplayskip{3pt}
\setlength\belowdisplayskip{3pt}
\begin{flalign}
&v^{*}=\arg \min\limits_{\textbf{v}\in{[0,1]^{n}}}\sum\limits_{i=1}^{n}v_i\cdot{L(\textbf{x}_{i},\textbf{w},b,y_i)}+f(\textbf{v},\lambda)
\end{flalign}}Here, $f(\lambda,v)$ is the self-paced regularization, and it satisfies the three following constraints \cite{Zhao:splfmf-AAAI2015}:
{\setlength\abovedisplayskip{3pt}
\setlength\belowdisplayskip{3pt}
\begin{itemize}
\setlength{\itemsep}{-5pt}
\setlength{\parsep}{-5pt}
\setlength{\parskip}{-5pt}
\item $f(\lambda,v)$ is convex with respect to $\emph{v}\in{[0,1]}$;\\
\item $v^*$ is monotonically increasing with respect to $\lambda$, and it holds that $\lim_{\lambda\rightarrow{0}}v^{*}=0$, and $\lim_{\lambda\rightarrow{\infty}}v^{*}\leq{1}$;\\
\item $v^*$ is monotonically decreasing with respect to $L(\textbf{x}_{i},\textbf{w},b,y_i)$, and it holds that $\lim_{L(\textbf{x}_{i},\textbf{w},b,y_i)\rightarrow{0}}v^{*}\leq{1}$, and $\lim_{L(\textbf{x}_{i},\textbf{w},b,y_i)}v^{*}=0$;
\end{itemize}}
From the three constraints mentioned above, we can easily find how the self-paced function works: by controlling the self-growth variable $\lambda$, SPL tends to select several easy examples (the loss is smaller) to learn first, and then gradually takes more, probably complex (the loss is larger), into the learning process. After all the instances are fed into the training model, SPL can get a more 'mature' model than the algorithm without self-paced learning scheme.

Learning from the two strategies above, we incorporate the SPL strategy into the PLL framework and propose the SP-PLL algorithm, which is introduced in the following section in a more detailed manner.

\section{The SP-PLL Approach}
Here, we first introduce how we integrate the self-paced scheme into the task of PLL, and then present the formulation of SP-PLL. After that, we give an efficient algorithm to solve the optimization problem of our proposed method.
\subsection{Formulation}
As is shown in Section \ref{Sec:background_partial_label}, compared with other methods based on margin criterion, M3PL is an effective algorithm to alleviate the noise produced by the false labels in candidate label sets. Nonetheless, during the optimization iterations, M3PL ignores the fact that each training instance and the corresponding candidate labels, which have varying complexity, often contribute differently to the learning result. And the instances assigned with false candidate labels will damage the effectiveness and robustness of the learning model.

To overcome the potential shortcoming mentioned above, self-paced scheme is incorporated into our framework, which has the following advantages: 1) it can avoid the negative effect produced by the instances assigned with unreliable label 2) and it can make the instances assigned with high-confidence labels contribute more to the learning model. Specifically, during each learning iteration, we fix the assigned labels $\textbf{y}=\{y_1,y_2,\ldots,y_n\}$ to update the classifier $\bm{\Theta}$. The instances assigned with high-confidence labels (i.e. the loss is smaller) can be learned first, and then the instances assigned with low-confidence labels (i.e. the loss is larger) can be admitted into the learning process, when the model has already become mature and the unreliable labels associated with the untrained instances have also become more reliable.

Following the proposed method, we design SP-PLL according to the steps as follows:
Firstly, we define the loss $L(\textbf{x}_{i},\textbf{w},b,y_i)$ by deforming the hinge loss, which is defined as:{
\setlength\abovedisplayskip{3pt}
\setlength\belowdisplayskip{3pt}
\begin{flalign}
L(\textbf{x}_{i},\textbf{w},b,y_i)=
\begin{cases}
0,&L(\textbf{x}_{i},\textbf{w},b,y_i)\leq{0}\\
1-{\xi}_i,&0<L(\textbf{x}_{i},\textbf{w},b,y_i)<1\\
1,&L(\textbf{x}_{i},\textbf{w},b,y_i)\geq{1}\\
\end{cases}
\label{Eq:lossfunction}
\end{flalign}}where, ${\xi}_i=[(\textbf{w}_{y_i}^{\top}\cdot\textbf{x}_i+b_{y_i})-\max_{\tilde{y_i}\not= y_i}(\textbf{w}_{\tilde{y_i}}^{\top}\cdot\textbf{x}_i+b_{\tilde{y_i}})]$, which is the margin between modeling outputs from ground-truth label and that from all other labels.

Next, we choose a suitable self-paced regularizer $f(\textbf{v},\lambda)$ for SP-PLL framework, which is associated with the weight values of the training examples. Learning from \cite{Zhao:splfmf-AAAI2015}, soft weighting can assign real-valued weights, which tends to reflect the latent importance of training samples more faithfully. Meanwhile, soft weighting has also been demonstrated that it is more effective than hard weighting in many real applications. Thus, we choose the soft SP-regularizer as follows:
{\setlength\abovedisplayskip{3pt}
\setlength\belowdisplayskip{3pt}
\begin{flalign}
f(\textbf{v},\lambda)=\sum\limits_{i=1}^{n}\frac{\lambda}{2}\cdot({v_i}^{2}-2{v}_i)
\label{Eq:regularizer}
\end{flalign}}

Finally, we integrate the loss Eq.(\ref{Eq:lossfunction}) and SP-regularizer Eq.(\ref{Eq:regularizer}) into the partial-label learning framework, and then get the framework of SP-PLL corresponds to the following optimization problem \textbf{OP(2)}:
{\setlength\abovedisplayskip{10pt}
\setlength\belowdisplayskip{10pt}
\begin{small}
\begin{flalign*}
&\min\limits_{\substack{\textbf{w},b\\0\leq{y_i}\leq{q}\\\textbf{v}\in{[0,1]^{n\times{1}}}}} \!\!\!\!\mathcal{C}\sum\limits_{i=1}^{n}{v_i}\!\cdot\!{L(\textbf{x}_{i},\textbf{w},b,y_i)}\!+
\!\frac{1}{2}\sum\limits_{p=1}^{q}\|\textbf{w}_{p}\|^{2}\!+\!\sum\limits_{i=1}^{n}\frac{\lambda}{2}\cdot({v_i}^{2}\!-\!2{v}_i)\\
&s.t.:
\begin{cases}
L(\textbf{x}_{i},\textbf{w},b,y_i)\!=\!1\!-\![(\textbf{w}_{y_i}^{\top}\!\cdot\!\textbf{x}_i\!+\!b_{y_i})\!-\!\max\limits_{\tilde{y_i}\not= y_i}(\textbf{w}_{\tilde{y_i}}^{\top}\!\cdot\!\textbf{x}_i\!+\!b_{\tilde{y_i}})]\\
L(\textbf{x}_{i},\textbf{w},b,y_i){\geq}{0},  \forall{i\in\{1,2,\ldots,n\}}\\
\textbf{y}\in\mathcal{S}\\
\sum_{i=1}^{n}\mathbb{I}(y_i=p)=n_p,  \forall{p\in\{1,2,\ldots,q\}}\\
\end{cases}
\end{flalign*}
\end{small}}where $n_p$ is the prior number of training instances belonging to the \emph{p}-th class label in $\mathcal{Y}$ and $\sum_{p=1}^{q}n_p=m$. The $n_p$ is defined as:{
\setlength\abovedisplayskip{3pt}
\setlength\belowdisplayskip{3pt}
\begin{flalign}
n_p=
\begin{cases}
\lfloor\hat{n_p}\rfloor+1,& p\in\{1,2,\ldots,r\}, and\quad{n_r}>\hat{n_p}\\
\lfloor\hat{n_p}\rfloor,& otherwise
\end{cases}
\end{flalign}}here,$\lfloor\hat{n}_p\rfloor$ is the integer part of $\hat{n}_p$ and $\hat{n}_p=\sum_{i=1}^{m}\mathbb{I}(p\in{S}_i)\cdot\frac{1}{|S_i|}$, $r$ is the residual number after the rounding operation and $r=m-\sum_{p=1}^{q}\lfloor\hat{n}_p\rfloor$.

Since \textbf{OP(2)} is a optimization problem with mixed-type variables, alternating optimization procedure is a good choice to solve the problem. We give the optimization details in the following subsection.

\begin{algorithm}[tb]
\caption{The Algorithm of \textbf{SP-PLL}}
\begin{algorithmic}
\label{Algorithm one}
   \STATE {\bfseries Inputs:}\\
   \quad $\mathcal{D}$:the partial label training set,$\{(\textbf{x}_i,S_i)\}$;\\
   \quad $\textbf{v}$: the weight vector,$\textbf{v}=\{v_1,v_2,\ldots,v_n\}$;\\
   \quad $\mathcal{C}_{max}$: the maximum value of regularization parameter;\\
   \quad $\lambda$: the learning parameter;\\
   \quad $\textbf{x}^{*}$: the unseen instance;\\
   \STATE {\bfseries Process:}\\
   \STATE \textbf{1.} Initialize the weight vector $\textbf{v}$, the Specified value $\mu$;\\
   \STATE \textbf{2.} Initialize the regularization parameter $\mathcal{C}$ and $\lambda$;\\
   \STATE \textbf{3.} Initialize the coefficient matrix $\textbf{C}$, where $c_{pi}=\frac{1}{|S_i|}$ if $p\in{S_i}$, otherwise $c_{pi}=M$;\\
   \STATE \textbf{4.} Initialize the ground-truth label \textbf{y} according to \textbf{OP(5)};\\
   \STATE \textbf{5.} \textbf{while} $\mathcal{C}<\mathcal{C}_{max}$\\
   \STATE \textbf{6.} \quad $\mathcal{C}=\min{\{(1+\Delta)\mathcal{C},\mathcal{C}_{max}\}}$\\
   \STATE \textbf{7.} \quad \textbf{while} $\lambda > loss_{max}$\\
   \STATE \textbf{8.} \quad\quad Initialize the function value \emph{OFV} in \textbf{OP(2)};\\
   \STATE \textbf{9.} \quad\quad \textbf{repeat} \\
   \STATE \textbf{10.} \quad\quad\quad $\emph{OFV}_{old}=\emph{OFV}$;\\
   \STATE \textbf{11.} \quad\quad\quad Solve \textbf{OP(3)}, update \textbf{w},b;\\
   \STATE \textbf{12.} \quad\quad\quad update coefficient matrix \textbf{C};\\
   \STATE \textbf{13.} \quad\quad\quad update \textbf{y} according to \textbf{OP(5)};\\
   \STATE \textbf{14.} \quad\quad\quad Calculate the function value \emph{OFV} in \textbf{OP(2)};\\
   \STATE \textbf{15.} \quad\quad \textbf{until} $\emph{OFV}_{old}-\emph{OFV}<\delta$;\\
   \STATE \textbf{16.} \quad\quad update \textbf{v} according to Eq.(\ref{Eq:five});\\
   \STATE \textbf{17.} \quad\quad update $\lambda=\mu\cdot\lambda$
   \STATE \textbf{18.} \quad \textbf{end while};\\
   \STATE \textbf{19.} \textbf{end while};\\
   \STATE \textbf{20.} \textbf{return} $y^{*}=\arg\max_{p\in{\mathcal{Y}}}\textbf{w}_p^{\top}\cdot\textbf{x}^{*}+b_p$;\\
   \STATE {\bfseries Output:}\\
   \quad $y^{*}$: the predicted label for $\textbf{x}^{*}$;\\
\end{algorithmic}
\end{algorithm}
%

\subsection{Optimization}
During the process of alternating optimization, we first improve the optimization algorithm of \cite{Yu:mmpll-ML2015} to make it suitable for our method to update the variables $\textbf{y}$ and $\bm\Theta(\textbf{w},b)$ , which is briefly introduced in Part (A) and (B), respectively. Then, we give the optimization algorithm of self-paced learning to update $\textbf{v}$ that is used to control the weight of different instances in each iterative process, which is introduced in part (C). And finally we summarize the procedure of SP-PLL at the end of the section.
\subsubsection{\textbf{Update w,b with other variables fixed}}
After initializing the weight vector $\textbf{v}$ and the ground-truth labels $\textbf{y}$ of training examples, \textbf{OP(2)} turns to be the following optimization problem \textbf{OP(3)}:
{
\setlength\abovedisplayskip{5pt}
\setlength\belowdisplayskip{5pt}
\begin{flalign*}
&\min\limits_{\substack{\textbf{w},b}}\quad \mathcal{C}\sum\limits_{i=1}^{n}{v_i}\cdot{L(\textbf{x}_{i},\textbf{w},b,y_i)}+
\frac{1}{2}\sum\limits_{p=1}^{q}\|\textbf{w}_{p}\|^{2}\\
&s.t.:
\begin{cases}
L(\textbf{x}_{i},\textbf{w},b,y_i)\!=\!1\!-\![(\textbf{w}_{y_i}^{\top}\!\cdot\!\textbf{x}_i\!+\!b_{y_i})\!-\!\max\limits_{\tilde{y_i}\not= y_i}(\textbf{w}_{\tilde{y_i}}^{\top}\!\cdot\!\textbf{x}_i\!+\!b_{\tilde{y_i}})]\\
L(\textbf{x}_{i},\textbf{w},b,y_i){\geq}{0},\quad  \forall{i\in\{1,2,\ldots,n\}}
\end{cases}
\end{flalign*}}

As is described in \textbf{OP(3)}, fixing the variables \textbf{y}, \textbf{(OP3)} is a typical single-label multi-class maximum margin optimization problem \cite{Crammer:Otaiomkbvm-JMLR2013}\cite{Hsu:acomfmsvm-IEEET2002}, which can be solved by utilizing the multi-class SVM implementations, such as liblinear toolbox\cite{Fan:liblinear-JMLR2008}.
\subsubsection{\textbf{Update y with other variables fixed}}
By fixing the classification model $\bm{\Theta}=\{\textbf{w},b\}$ and the weight vector $\textbf{V}$, \textbf{OP(2)} can turn to be the following optimization problem \textbf{OP(4)}:
{
\setlength\abovedisplayskip{1pt}
\setlength\belowdisplayskip{1pt}
\begin{flalign*}
&\min\limits_{\substack{\textbf{y}}} \quad \sum\limits_{i=1}^{n}v_{i}\cdot{L(\textbf{x}_{i},\textbf{w},b,y_i)}\\
&s.t.\!
\begin{cases}
L(\textbf{x}_{i},\textbf{w},b,y_i)\!=\!1\!-\![(\textbf{w}_{y_i}^{\top}\!\!\cdot\!\textbf{x}_i\!+\!b_{y_i})\!-\!\max\limits_{\tilde{y_i}\not= y_i}(\textbf{w}_{\tilde{y_i}}^{\top}\!\!\cdot\!\textbf{x}_i\!+\!b_{\tilde{y_i}})]\\
L(\textbf{x}_{i},\textbf{w},b,y_i){\geq}{0},\quad  \forall{i\in\{1,2,\ldots,n\}}\\
\textbf{y}\in\mathcal{S}\\
\sum_{i=1}^{n}\mathbb{I}(y_i=p)=n_p,\quad \forall{p\in\{1,2,\ldots,q\}}\\
\end{cases}
\end{flalign*}
}

To simplify the \textbf{OP(4)}, inspired by \cite{Yu:mmpll-ML2015}, we first replace $L(\textbf{x}_{i},\textbf{w},b,y_i)$ with $ \{\max(0,1-[(\textbf{w}_{y_i}^{\top}\!\!\cdot\!\textbf{x}_i+b_{y_i})-\max_{\tilde{y_i}\not= y_i}(\textbf{w}_{\tilde{y_i}}^{\top}\!\!\cdot\!\textbf{x}_i+b_{\tilde{y_i}})])\}$ according to the first two constraints. Then, we define a labeling matrix $\textbf{Z}=[z_{pi}]_{q\times{n}}$ and a coefficient matrix $\textbf{C}=[c_{pi}]_{q\times{n}}$, where $z_{pi}=1$ indicates that the ground-truth label of $\textbf{x}_i$ belongs to the $\emph{p}$-th class and $c_{pi}$ represents the loss that the $\emph{p}$-th class label is assigned to the candidate examples $\textbf{x}_i$. Here, if $y_p\in{S_i}$, $c_{pi}=\max{(0,L(\textbf{x}_{i},\textbf{w},b,y_p))}$, otherwise $c_{pi}$ would obtain a large value. Based on the steps mentioned above, \textbf{OP(4)} can be formulated as the following optimization problem \textbf{OP(5)}:
{
\setlength\abovedisplayskip{5pt}
\setlength\belowdisplayskip{5pt}
\begin{flalign*}
&\min\limits_{\textbf{Z}} \quad \sum\limits_{p=1}^{q}\sum\limits_{i=1}^{n}{{v_i}\cdot{c_{pi}}\cdot{z_{pi}}}\\
&s.t.:
\begin{cases}
\sum_{p=1}^{q}{z}_{pi}=1,\quad  \forall{i\in\{1,2,\ldots,n\}}\\
\sum_{i=1}^{n}{z}_{pi}={n}_{p},\quad \forall{p\in\{1,2,\ldots,q\}}\\
{z}_{pi}\in{[0,1]}
\end{cases}
\end{flalign*}
}\textbf{OP(5)} is an easy linear programming problem, which can be solved by utilizing the standard LP solver.

\subsubsection{\textbf{Update v with other variables fixed}}
By fixing the classification model $\textbf{w},b$, and the ground-truth labels $\textbf{y}$, we update the weight vector $\textbf{v}$ by solving the following optimization problem \textbf{OP(6)}:
{
\setlength\abovedisplayskip{5pt}
\setlength\belowdisplayskip{5pt}
\begin{flalign*}
&\min\limits_{\textbf{v}} \mathcal{C}\sum\limits_{i=1}^{n}{v_i}\cdot{L(\textbf{x}_{i},\textbf{w},b,y_i)}+\sum\limits_{i=1}^{n}\frac{\lambda}{2}\cdot({v_i}^{2}-2{v}_i)
\end{flalign*}}

According to \textbf{OP(6)}, $v_i$ in the SP-PLL model could be computed as\\
\begin{flalign}
v^{*}(\lambda,L)=
\begin{cases}
\label{Eq:five}
1-\frac{L(\textbf{x}_{i},\textbf{w},b,y_i)}{\lambda},&L(\textbf{x}_{i},\textbf{w},b,y_i)\leq{\lambda}\\
0,&L(\textbf{x}_{i},\textbf{w},b,y_i)>{\lambda}
\end{cases}
\end{flalign}
here, it is easy to see that examples assigned with higher-confidence labels (i.e. $L(\textbf{x}_{i},\textbf{w},b,y_i)$ is smaller) can get higher weight values than the examples assigned with lower-confidence labels (i.e. $L(\textbf{x}_{i},\textbf{w},b,y_i)$ is larger), while examples assigned with extremely unreliable labels (i.e. $L(\textbf{x}_{i},\textbf{w},b,y_i)$ is larger than $\lambda$) even can not be chosen in previous iterative flow, which is the so called '\emph{learning from easy to hard}' self-paced scheme.

During the entire process of alternating optimization, we first initialize the required variables, and then repeat the above process until the algorithm converges. Finally, we get the predicted labels of the unseen instances according to the trained classifier. The detail process of SP-PLL is summarized in Algorithm \textbf{ \ref{Algorithm one}}.

\begin{figure*}[!ht]
\centering
\begin{tabular}{cc}
\includegraphics[width = 3in,height=2.2in]{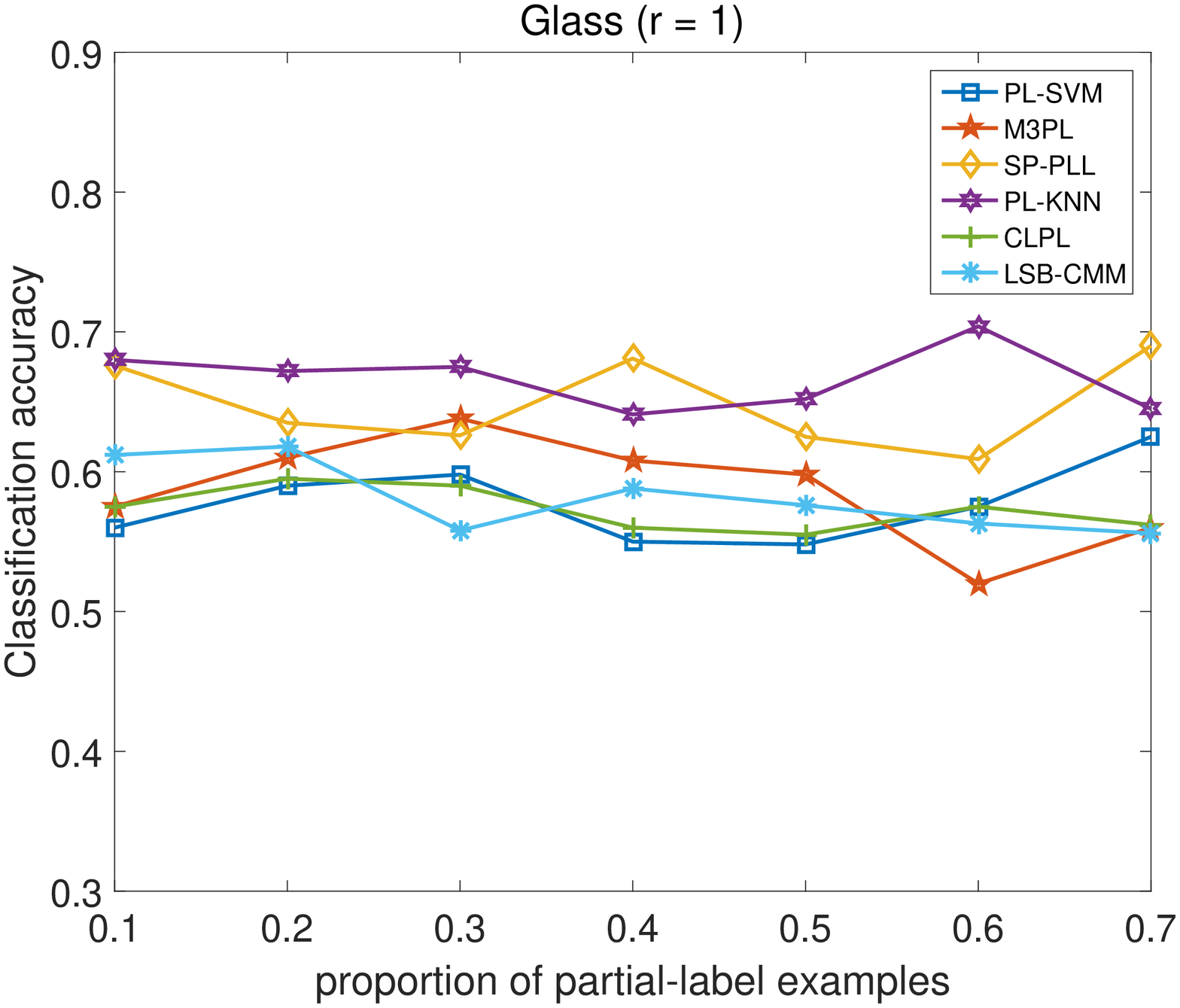}&\includegraphics[width = 3in,height=2.2in]{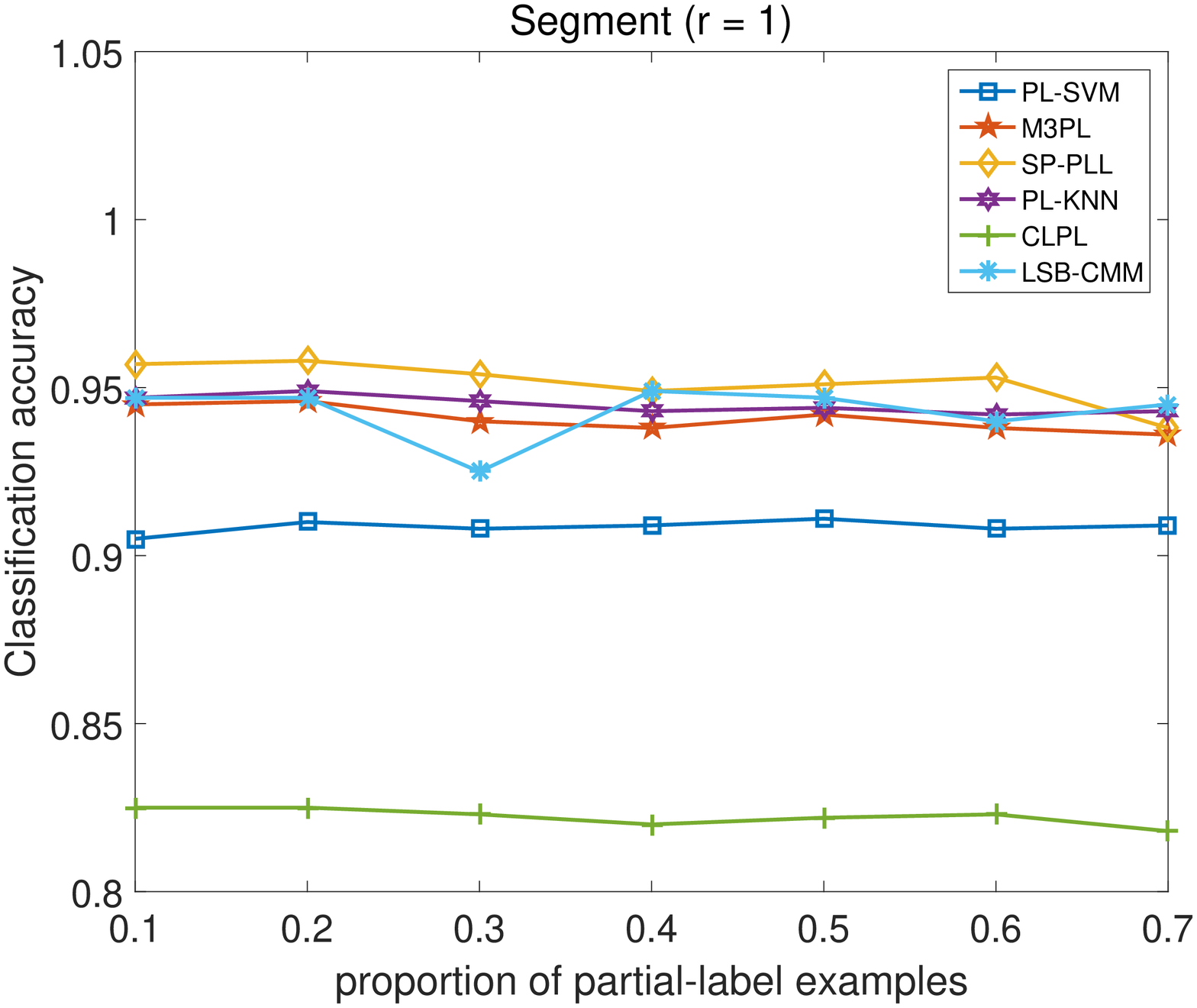}\\
\includegraphics[width = 3in,height=2.2in]{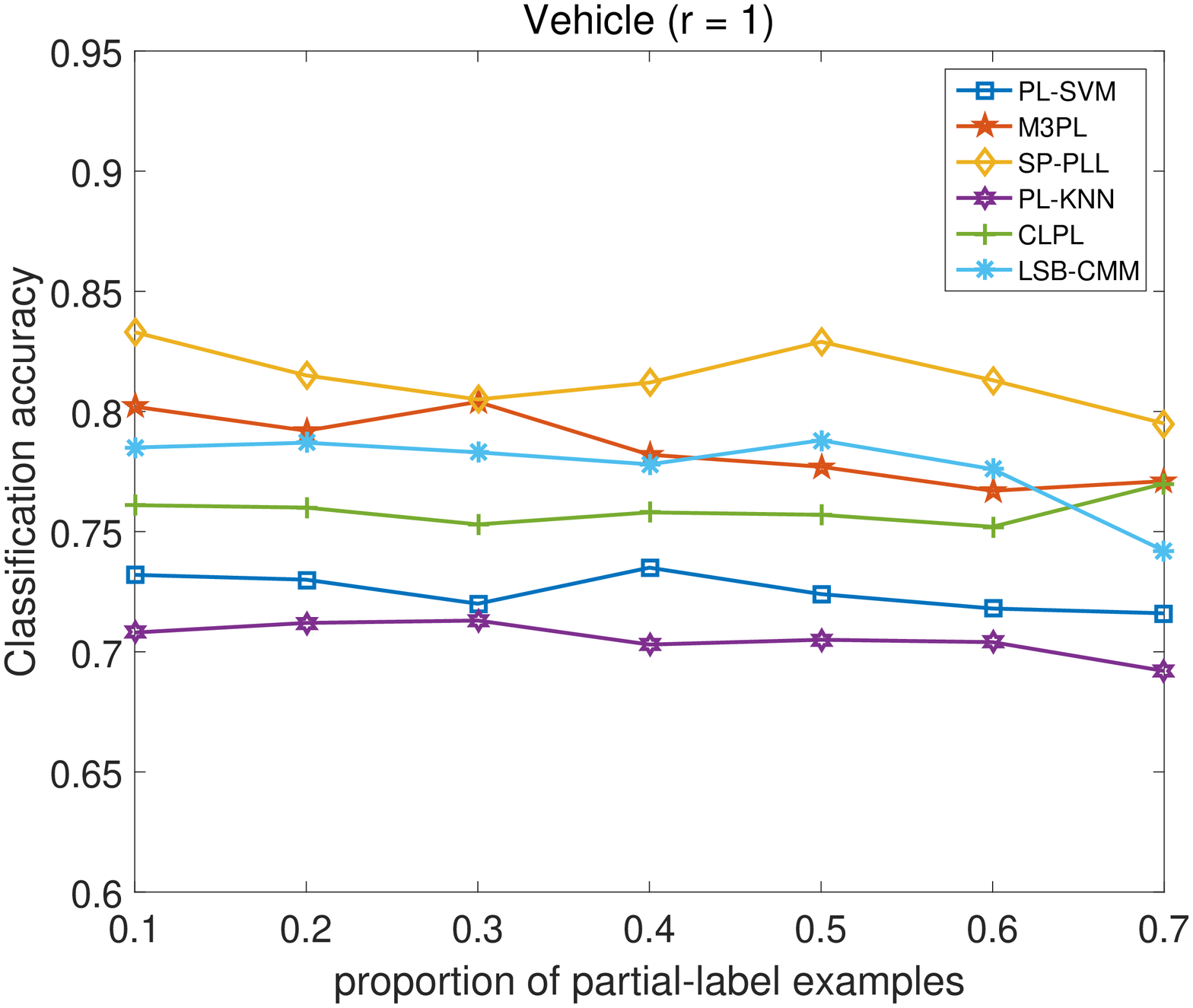}&\includegraphics[width = 3in,height=2.2in]{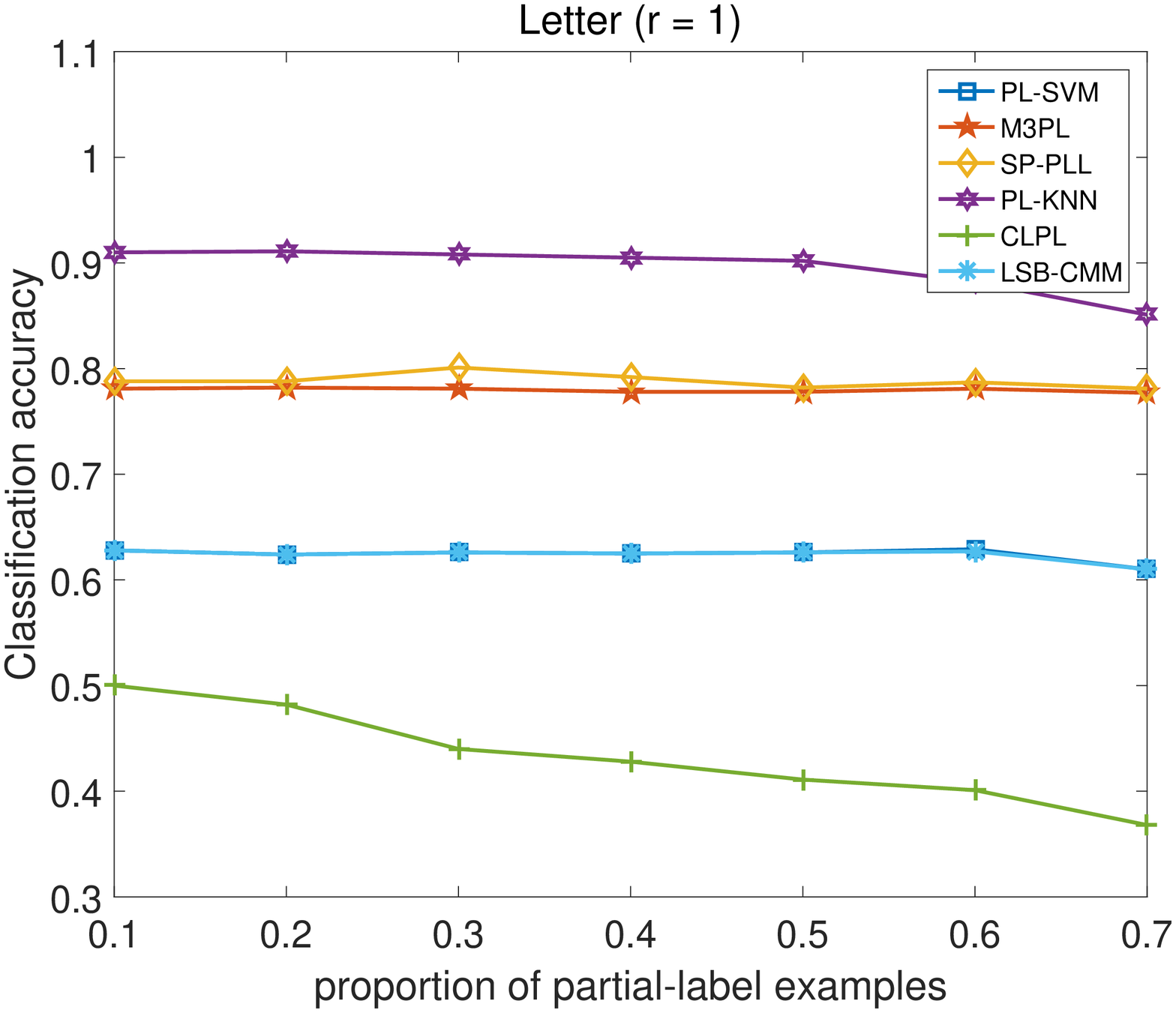}\\
\end{tabular}
\vspace{3mm}
\caption{The classification accuracy of several comparing methods on four controlled UCI data sets with one false positive candidate labels (r = 1)}
\label{Figure2}
\vspace{0mm}
\end{figure*}

\begin{figure*}[!ht]
\centering
\begin{tabular}{cc}
\includegraphics[width = 3in,height=2.2in]{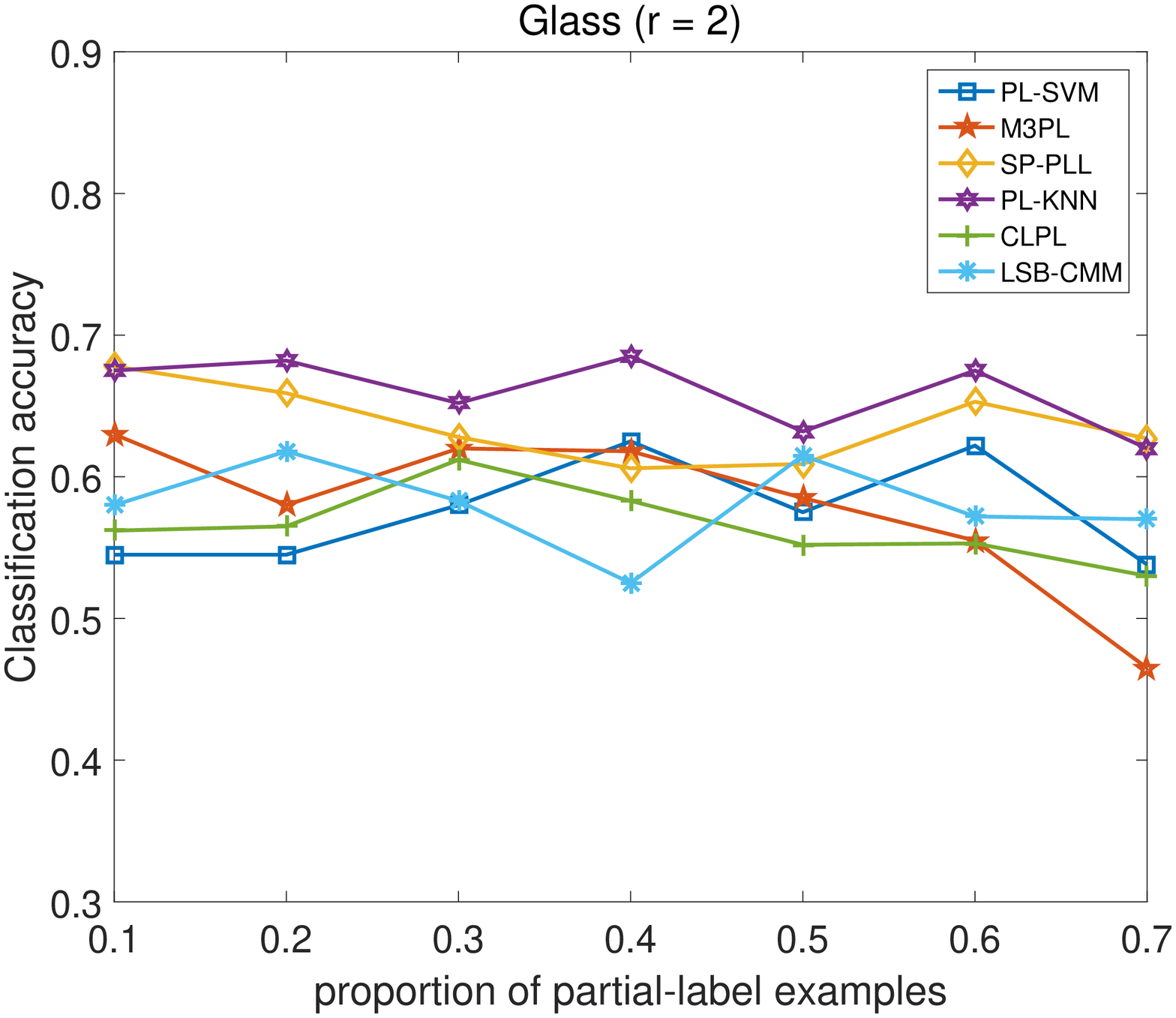}&\includegraphics[width = 3in,height=2.2in]{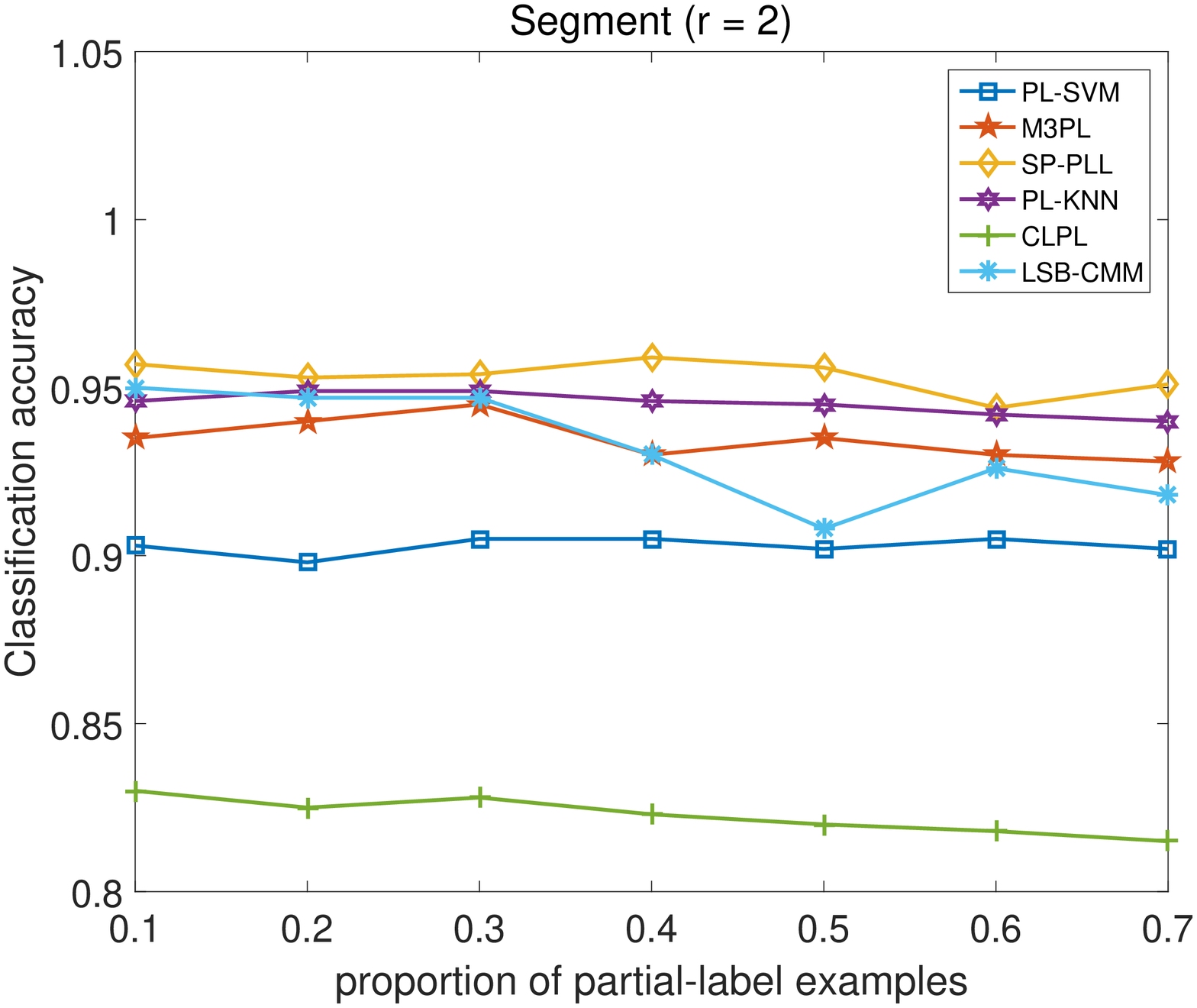}\\
\includegraphics[width = 3in,height=2.2in]{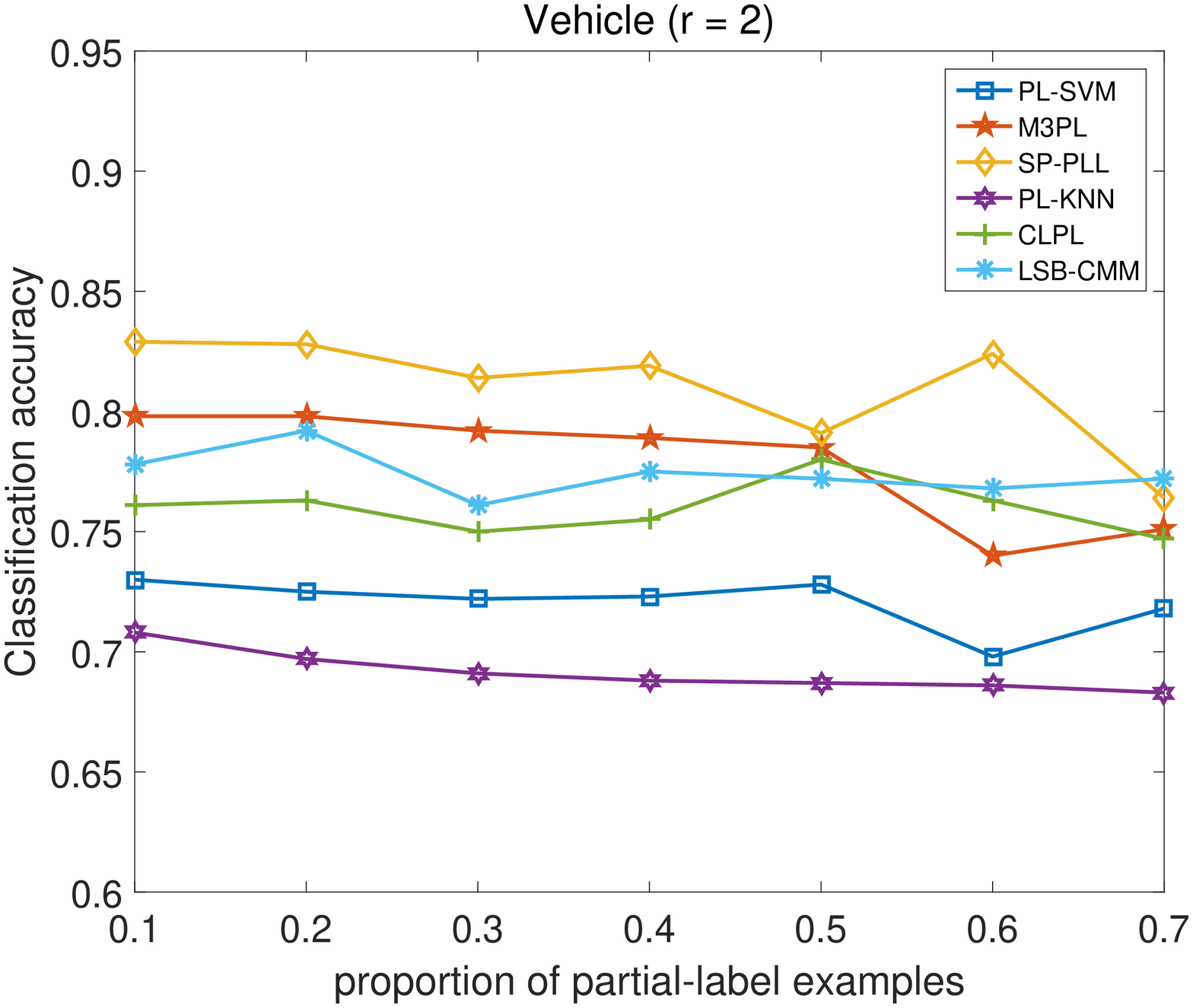}&\includegraphics[width = 3in,height=2.2in]{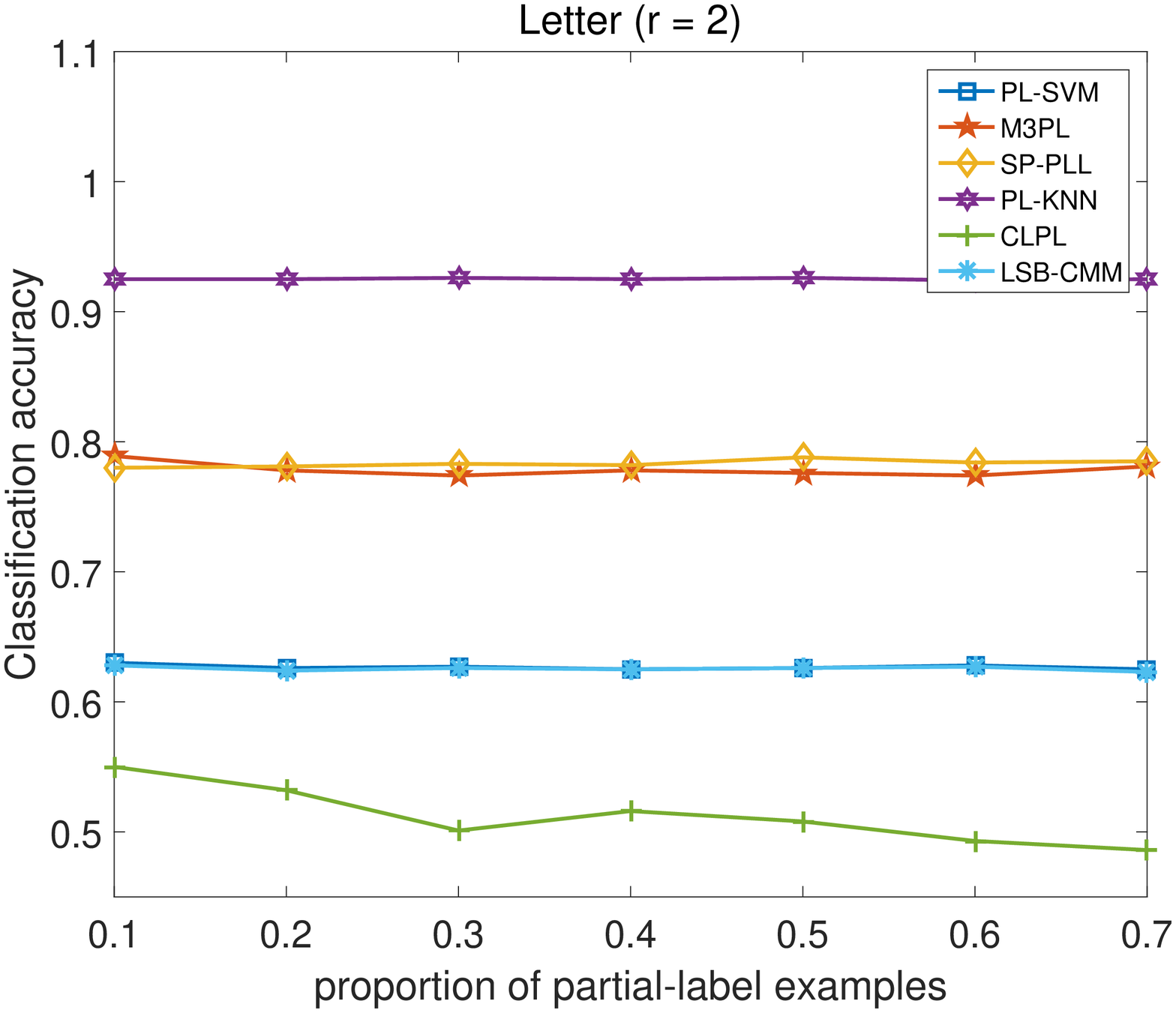}\\
\end{tabular}
\vspace{3mm}
\caption{The classification accuracy of several comparing methods on four controlled UCI data sets with one false positive candidate labels (r = 2)}
\label{Figure3}
\vspace{0mm}
\end{figure*}

\begin{table*}[!ht]
\centering
\caption{\textbf{Characteristics of the Experimental data sets}}
\label{table1}
\resizebox{!}{1.5cm}{
\begin{tabular}{cccccc}
\cline{1-5}
\hline \hline
Real World data sets   & EXP*     & FEA*     & CL*         & AVG-CL*   &TASK DOMAIN     \\ \hline
Lost          & 1122     & 108      & 16          & 2.33      &\emph{Automatic Face Naming} \cite{Cour:lfpl-JMLR2011}      \\
BirdSong     & 4998     & 38       & 13          & 2.18      &\emph{Bird Sound Classification} \cite{Liu:acmmmfsll-NIPS2012}      \\
MSRCv2       & 1758     & 48       & 23          & 3.16      &\emph{Image Classification} \cite{Briggs:rlsimfmia-KDDM2012}  \\
Soccer Player & 17472    & 279      & 171         & 2.09      &\emph{Automatic Face Naming} \cite{Guill:mimlfalbof-ECCV2010} \\
FG-NET        & 1002     & 262      & 99          & 7.48      &\emph{Facial Age Estimation} \cite{Panis:FG-NET-JAH2015} \\ \hline \hline
\end{tabular}}
\vspace{-1mm}
\end{table*}

\section{Experiments}
\subsection{Experimental Setup}
To evaluate the performance of the proposed SP-PLL algorithm, we implement experiments on two controlled UCI data sets and five real-world data sets: \textbf{(1) UCI data sets}. Under different configurations of two controlling parameters (i.e. $\emph{p}$ and $\emph{r}$), the two UCI data sets generate 84 partial-labeled data sets with different configurations \cite{Cour:lfpl-JMLR2011}\cite{Chen:allud-IEEET2014}. Here, $\emph{p}\!\in\!{\{0.1,0.2,\ldots,0.7\}}$ is the proportion of partial-labeled examples and $\emph{r}\!\in\!{\{1,2,3\}}$ is the number of candidate labels instead of the correct one. \textbf{(2) Real-World (RW) data sets }. These data sets are collected from the following task domains: (A) {\emph{Facial Age Estimation}}; (B) {\emph{Automatic Face Naming}}; (C) {\emph{Image Classification}}; (D) {\emph{Bird Sound Classification}};

\begin{table}[!ht]
\centering
\caption{\textbf{Characteristics of the Experimental data sets}}
\label{table2}
\begin{tabular}{cccc|c}
\cline{1-5}
\hline \hline
UCI data sets  & EXP*     & FEA*     & CL*         & Configurations           \\ \hline
glass         & 214      & 10       & 7           &          \\
segment       & 2310     & 18       & 7           & $r = 1, p\in\{0.1,\ldots,0.7\}$  \\
vehicle       & 846      & 18       & 4           & $r = 2, p\in\{0.1,\ldots,0.7\}$  \\
letter        & 5000     & 16       & 26          & $r = 3, p\in\{0.1,\ldots,0.7\}$                \\ \hline \hline
\end{tabular}
\vspace{-1mm}
\end{table}

Table \ref{table2} and Tabel \ref{table1} separately summarizes the characteristics of the above UCI data sets and real world data sets, including the number of examples (\textbf{EXP*}), the number of the feature (\textbf{FEA*}), and the whole number of class labels (\textbf{CL*}).

Meanwhile, we employ eight state-of-the-art partial label learning algorithms\footnote{We partially use the open source codes from Zhang Minling's homepage: http://cse.seu.edu.cn/PersonalPage/zhangml/} for comparative studies, where the configured parameters of each method are utilized according to that suggested in the respective literatures:

{
\begin{itemize}
\item \textbf{PL-SVM} \cite{Nguyen:cwpl-KDDM2008}: Based on the maximum-margin strategy, it gets the predicted-label according to calculating the maximum values of model outputs. [suggested configuration: $\lambda\!\in\!{\{10^{-3},10^{-2},\ldots,10^{3}\}}$] ;
\item \textbf{PL-KNN} \cite{Huller:lfale-LNCS2005}: Based on \emph{k}-nearest neighbor method, it gets the predicted-label according to averaging the outputs of the \emph{k}-nearest neighbors. [suggested configuration: k=10];
\item \textbf{CLPL} \cite{Cour:lfpl-JMLR2011}: A convex optimization partial-label learning method via averaging-based disambiguation [suggested configuration: SVM with hinge loss];
\item \textbf{LSB-CMM} \cite{Liu:acmmmfsll-NIPS2012}: Based on maximum-likelihood strategy, it gets the predicted-label according to calculating the maximum-likelihood value of the model with unseen instances input. [suggested configuration: \emph{q} mixture components];
\item \textbf{M3PL} \cite{Yu:mmpll-ML2015}: Originated from PL-SVM, it is also based on the maximum-margin strategy, and it gets the predicted-label according to calculating the maximum values of model outputs. [suggested configuration: $C_{max}\in{\{0.01,0.1,\ldots,100\}}$] ;
\item \textbf{PL-LEAF} \cite{Zhang:pllvfad-TKDD2016}: A partial-label learning method via feature-aware disambiguation [suggested configuration: k=10, $C_1=10$, $C_2=1$];
\item \textbf{IPAL} \cite{Zhang:stpllpaiba-IJCAI2015}: it disambiguates the candidate label set by utilizing instance-based techniques [suggested configuration: k=10];
\item \textbf{PL-ECOC} \cite{zhang:dfpll-IEEET2017}: Based on a coding-decoding procedure, it learns from partial-label training examples in a disambiguation-free manner [suggested configuration: the codeword length $L = \lceil\log_{2}(q)\rceil$];
\end{itemize}}
Inspired by \cite{Nguyen:cwpl-KDDM2008} and \cite{Yu:mmpll-ML2015}, we set $C_{max}$ among $\{0.01,\ldots,100\}$ via cross-validation. And the initial value of $\lambda$ is empirically set to more than $0.5$ to guarantee that at least half of the training instances can be learned during the first iterative optimization process. Furthermore, the other variables are set as $\Delta=0.5$, $loss_{max}=10^{-4}$, and $\mu=1.05$. After initializing the above variables, we adopt ten-fold cross-validation to train each data set and report the average classification accuracy on each data set.


\begin{figure*}[!ht]
\centering
\begin{tabular}{cc}
\includegraphics[width = 3in,height=2.2in]{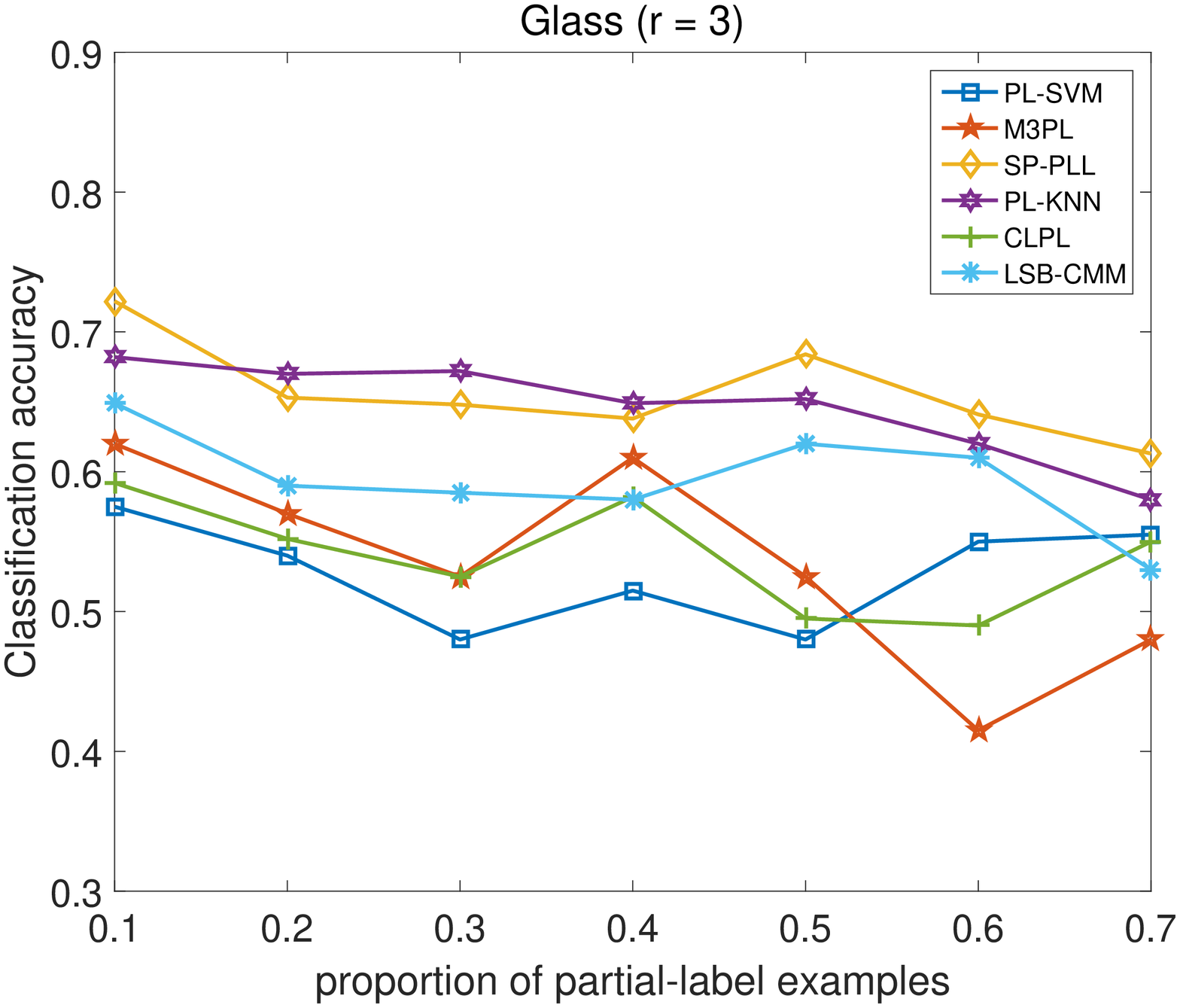}&\includegraphics[width = 3in,height=2.2in]{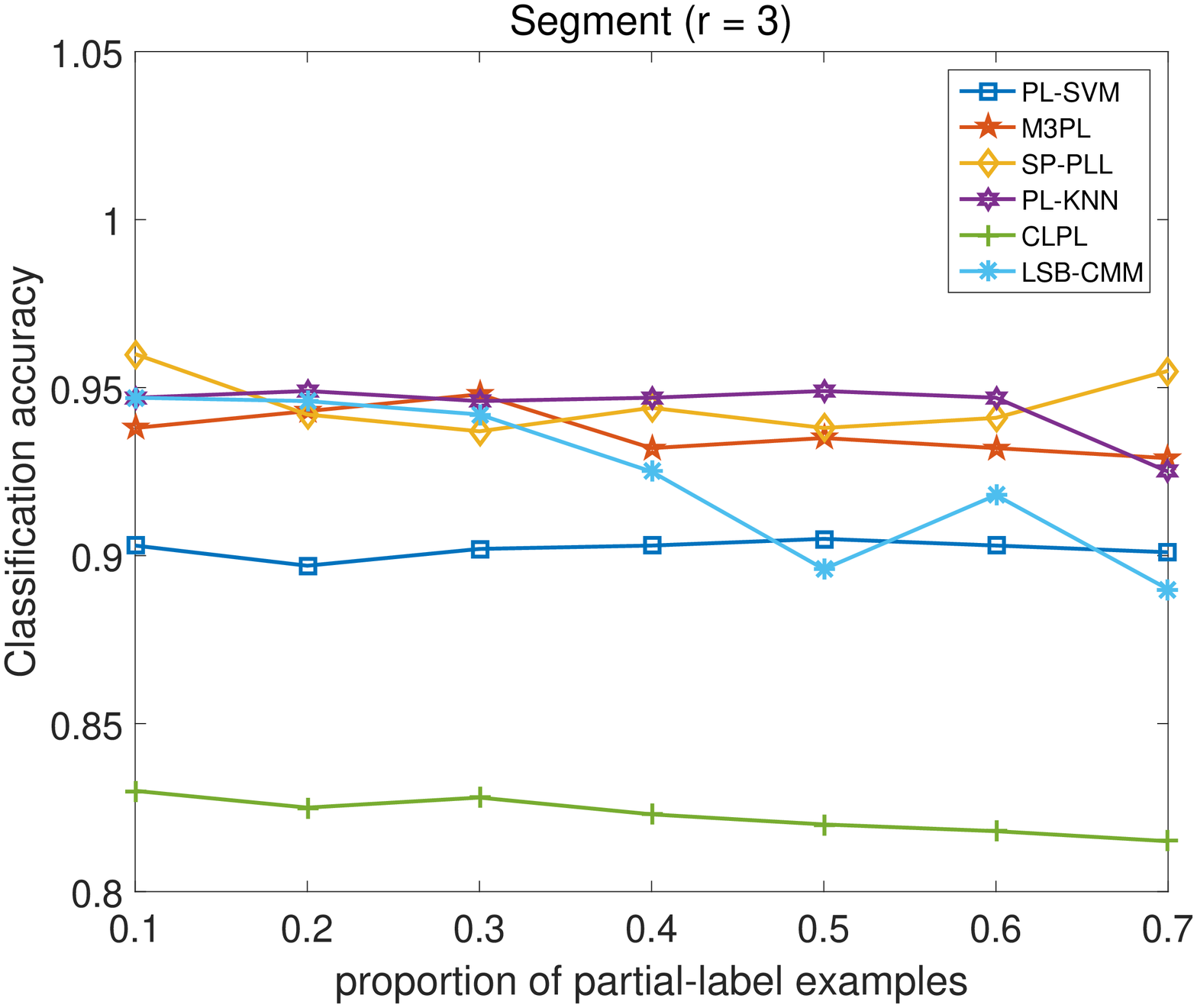}\\
\includegraphics[width = 3in,height=2.2in]{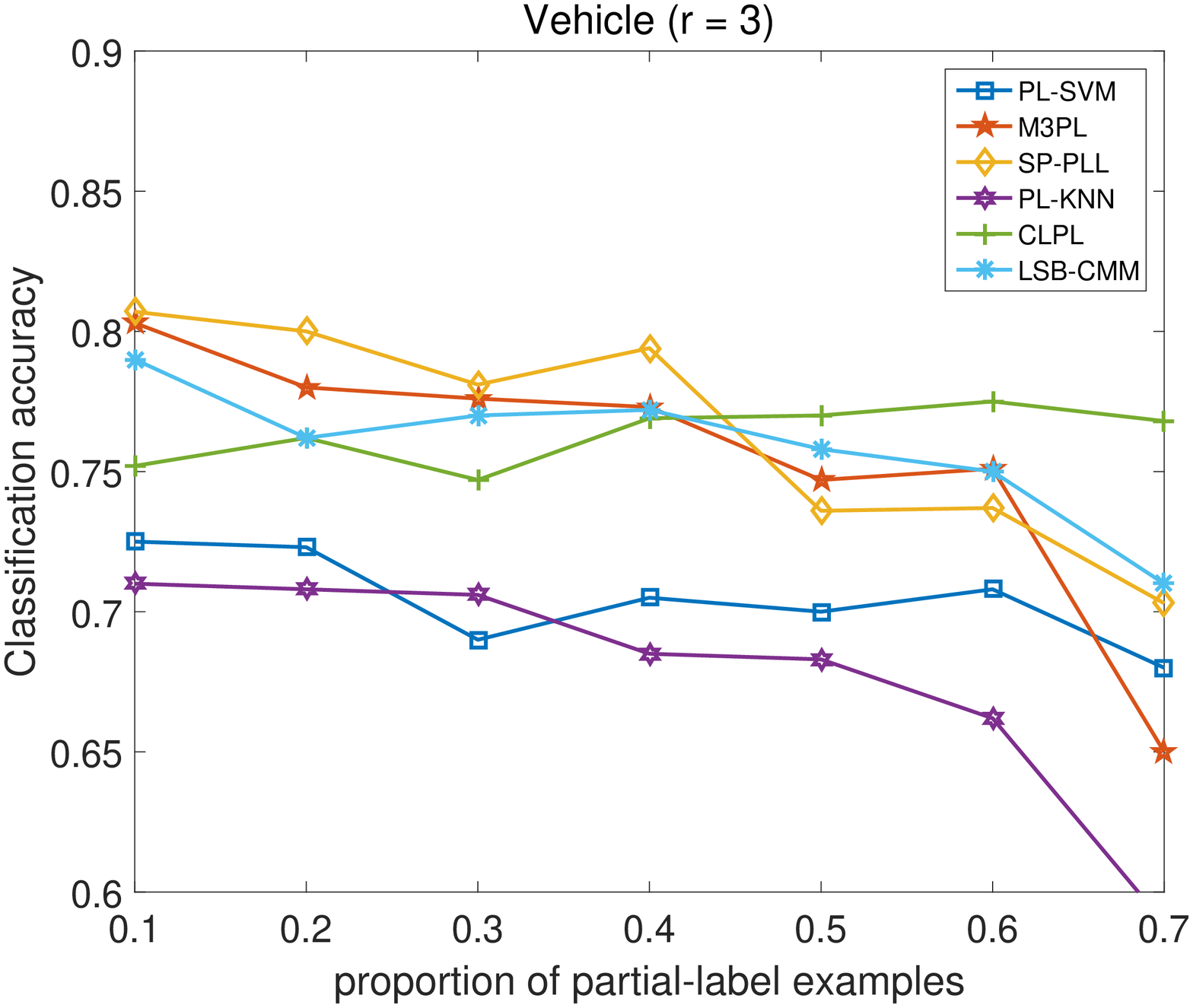}&\includegraphics[width = 3in,height=2.2in]{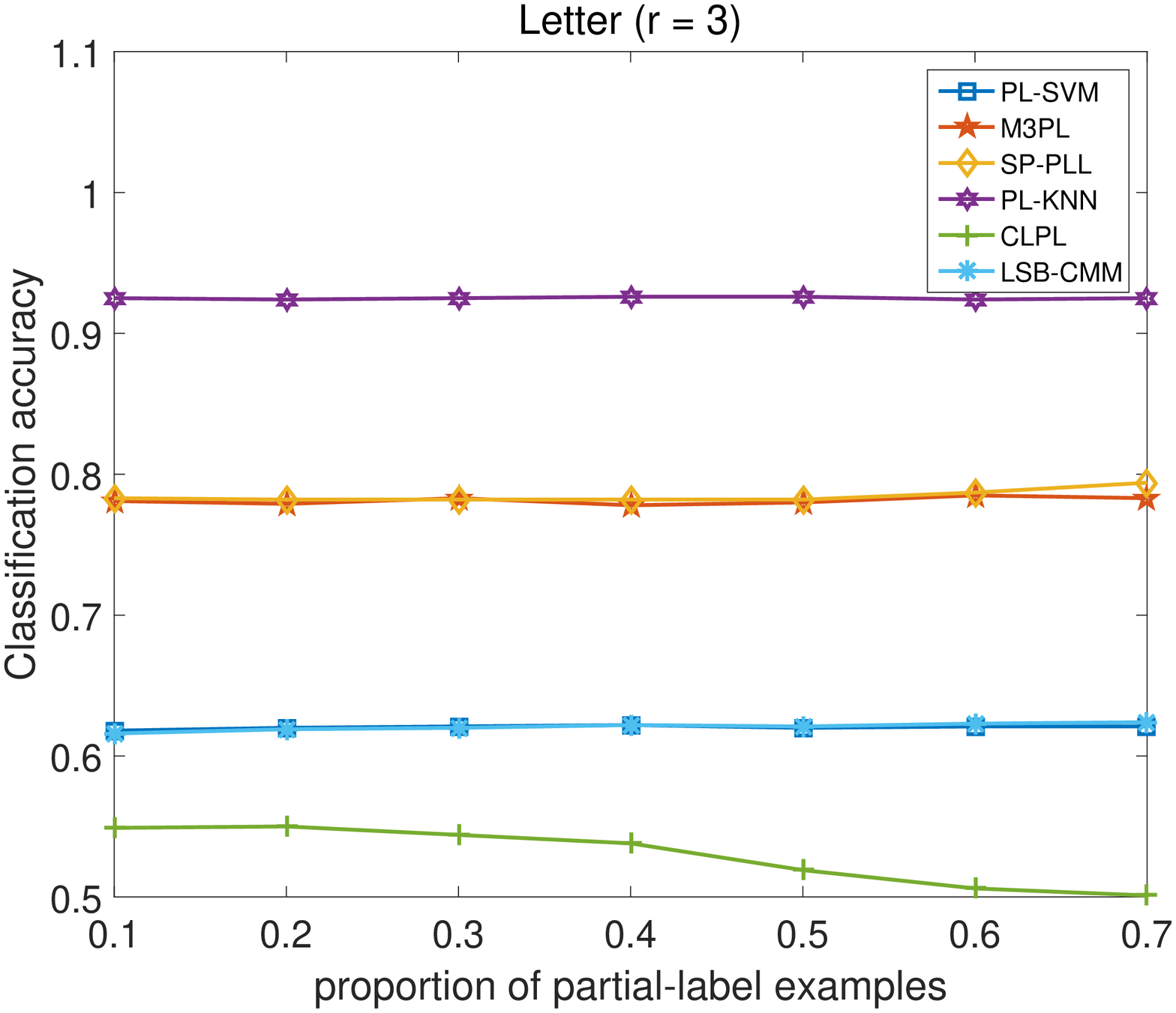}\\
\end{tabular}
\vspace{3mm}
\caption{The classification accuracy of several comparing methods on four controlled UCI data sets with one false positive candidate labels (r = 3)}
\label{Figure4}
\vspace{0mm}
\end{figure*}

\subsection{Experimental Results}
In our paper, the experimental results of the comparing algorithms originate from two aspects: one is from the results we implement by using the source codes provided by the authors; another is from the results shown in the respective literatures.
\subsubsection{\textbf{UCI data sets}}
We compare the SP-PLL with the comparing methods PL-SVM and M3PL, which SP-PLL originates from, to evaluate the effect of SP-regularizer in SP-PLL. Meanwhile, we also compare the proposed method (SP-PLL) with other baseline methods that are not based on maximum-margin strategy. The classification accuracy on the four UCI data sets ( \emph{glass}, \emph{segment}, \emph{vehicle} and \emph{letter}, each data set with $7\!\!\times\!\!3\!\!=\!\!21$ configurations) are shown in Figure 2.

\begin{itemize}
\item \emph{A)} SP-PLL achieves superior performance against M3PL in 95.24\% cases and PL-SVM in 97.62\% cases respectively (total cases: 3*7*4 = 84);
\item \emph{B)} SP-PLL outperforms PL-KNN in 60.72\% cases and is inferior to PL-KNN in 39.28\% cases;
\item \emph{C)} SP-PLL has been outperformed by CLPL in only 3 cases and by LSB-CMM in only 4 cases, and it outperforms them in the rest cases.
\end{itemize}

As is described in Figure \ref{Figure2}-\ref{Figure4}, SP-PLL achieves superior performance against the two algorithms (i.e. PL-SVM and M3PL) that our methods originate from and obtains competitive performance against than other comparing methods, which is embodied in the following aspects:
\begin{itemize}
\item \textbf{Average Classification Accuracy} With the increasing of \emph{p} (the proportion of partial-label examples) and \emph{r} (the number of extra labels in candidate label set), more noisy labels are added to the training data. As shown in Figure \ref{Figure2}, M3PL and PL-SVM are greatly influenced by these noises and the classification accuracy decreases significantly. In contrast, SP-PLL still performs well on disambiguating candidate labels, where the average classification accuracy of SP-PLL is $4.66\%$ higher than that of M3PL on the \emph{glass} data set, $0.71\%$ higher on the \emph{segment} data set, $0.53\%$ higher on the \emph{vehicle} data set and $0.49\%$ higher on the \emph{letter}, respectively.
\item \textbf{Max-Min and Standard deviation of Classification Accuracy} As more noisy candidate labels are gradually fed into the training data, the classification accuracy of M3PL declines dramatically. For the \emph{glass} data set, Max-Min and standard deviation of M3PL's classification accuracy separately reach to $20\%$ and 0.055, while SP-PLL reaches to only $10\%$ and 0.027. For \emph{segment} data set, the classification accuracy of SP-PLL and M3PL have the similar Max-Min value but the standard deviation of SP-PLL's classification accuracy is 0.002 smaller than that of M3PL's. And for vehicle data set, the classification accuracy of SP-PLL and M3PL have similar standard deviation value while the Max-Min of SP-PLL's classification accuracy is 0.024 smaller than that of M3PL's.  The above results demonstrate that the proposed SP-PLL is more robust than M3PL.
\item \textbf{Data Sets with Varying Complexities} According to the statistical comparisons of classification accuracy on the two data sets (accuracy on \emph{glass} is lower than \emph{segment}), we can see that examples in \emph{glass} is much more difficult to be disambiguated than that in \emph{segment}. However, SP-PLL can express more effective disambiguation ability on such difficult data set. Specifically, the performance on \emph{glass} is $7\%$ higher than that on \emph{segment}, which again demonstrates the disambiguation ability of the proposed SP-PLL.
\end{itemize}

\begin{table*}[!ht]
\centering
\setlength{\abovecaptionskip}{0pt}%
\setlength{\belowcaptionskip}{5pt}%
\caption{ Classification accuracy of each algorithm on real-world data sets. $\bullet/\circ$ indicates that SP-PLL is statistically superior / inferior to the algorithm. }
\label{Table3}
\vspace{2mm}
\resizebox{!}{2.5cm}{
\begin{tabular}{cccccc}
\toprule[0.5pt]
\toprule[0.5pt]
 & Lost  & MSRCv2 & BirdSong & SoccerPlayer & FG-NET
 \vspace{1mm} \\
\toprule[0.5pt]
SP-PLL  & \textbf{0.749$\pm$0.033}   & \textbf{0.581$\pm$0.010}    & 0.710$\pm$0.008    & 0.470$\pm$0.010          & \textbf{0.078$\pm$0.022}   \\
\toprule[1.0pt]
PL-SVM  & 0.639$\pm$0.056 $\bullet$ & 0.417$\pm$0.027 $\bullet$  & 0.671$\pm$0.018 $\bullet$    & 0.430$\pm$0.004 $\bullet$        & 0.058$\pm$0.010 $\bullet$ \\
M3PL    & 0.732$\pm$0.035 $\bullet$ & 0.546$\pm$0.030 $\bullet$  & 0.709$\pm$0.010 $\bullet$    & 0.446$\pm$0.013 $\bullet$        & 0.037$\pm$0.025 $\bullet$  \\
\toprule[1.0pt]
CLPL    & 0.670$\pm$0.024 $\bullet$ & 0.375$\pm$0.020 $\bullet$  & 0.624$\pm$0.009 $\bullet$    & 0.347$\pm$0.004 $\circ$        & 0.047$\pm$0.017 $\bullet$  \\
PL-KNN  & 0.332$\pm$0.030 $\bullet$ & 0.417$\pm$0.012 $\bullet$  & 0.637$\pm$0.009 $\bullet$    & 0.494$\pm$0.004 $\circ$       & 0.037$\pm$0.008 $\bullet$  \\
LSB-CMM & 0.591$\pm$0.019 $\bullet$ & 0.431$\pm$0.008 $\bullet$  & 0.692$\pm$0.015 $\bullet$    & 0.506$\pm$0.006 $\circ$        & 0.056$\pm$0.008 $\bullet$  \\
PL-LEAF & 0.664$\pm$0.020 $\bullet$ & 0.459$\pm$0.013 $\bullet$  & 0.706$\pm$0.012 $\bullet$    & 0.515$\pm$0.004 $\circ$        & 0.072$\pm$0.010 $\bullet$  \\
IPAL    & 0.726$\pm$0.041 $\bullet$ & 0.523$\pm$0.025 $\bullet$  & 0.708$\pm$0.014 $\bullet$    & \textbf{0.547$\pm$0.014} $\circ$        & 0.057$\pm$0.023 $\bullet$  \\
PL-ECOC & 0.703$\pm$0.052 $\bullet$ & 0.505$\pm$0.027 $\bullet$  & \textbf{0.740$\pm$0.016} $\circ$    & 0.537$\pm$0.020 $\circ$        &   0.040$\pm$0.018  $\bullet$     \\
\toprule[0.5pt]
\toprule[0.5pt]
\end{tabular}}
\vspace{-2mm}
\end{table*}

Besides, we note that the performance of SP-PLL is lower than PL-KNN on a few UCI data set (\emph{glass} and \emph{letter}). We attribute it to the difference of what the two methods are based on, where the former is based on maximum margin strategy and the latter is based on \emph{k}-NN strategy. For different data sets, varying learning strategies have difference in the performance of learning results. However, according to the above comparing results, our method (SP-PLL) not only outperforms all existing maximum-margin PLL methods but also obtains competitive performance as compared with most methods based on other strategies.

\subsubsection{\textbf{Real-world (RW) data sets}}
We compare the SP-PLL with all above comparing algorithms on the real-world data sets, and the comparison results are reported in Table \ref{Table3}, where the recorded results are based on ten-fold cross-validation.

It is easy to conclude that SP-PLL performs better than most comparing partial-label learning algorithms on these RW data sets. The superiority of SP-PLL can be embodied in the following two aspects:
\begin{itemize}
\item Compared with PL-SVM and M3PL, which are also based on maximum margin strategy, SP-PLL outperforms all of them on the whole RW data sets. Especially, the classification accuracy of the proposed method is 7\% more than M3PL's and 20\% more than PL-SVM's respectively on MSRCv2 data set; And on the FG-NET data set, SP-PLL achieves around 100\% classification accuracy improvement than M3PL.
\item SP-PLL also performs great superiority on some data sets when comparing with the other baseline algorithms. Specifically, for the same data set (such as \emph{Lost} and \emph{SoccerPlayer}), in contrast to the M3PL algorithm which has worse performance than the other algorithms, the proposed SP-PLL algorithm usually performs well, which again demonstrates the advantage of incorporating SPL regime in our proposed method.
\end{itemize}


\begin{figure*}[!ht]
\centering
\begin{tabular}{cc}
\includegraphics[width = 3in,height=2.2in]{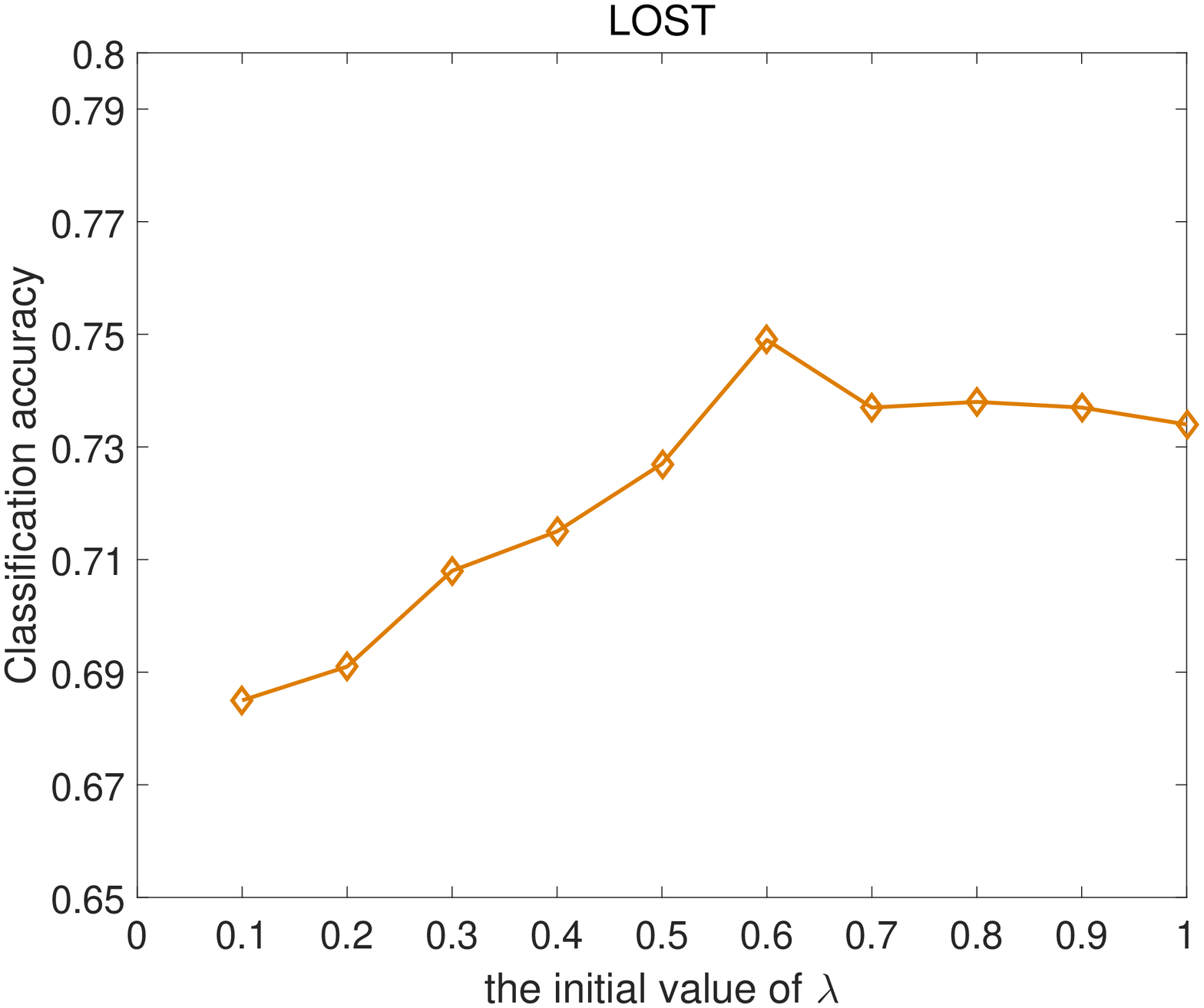}&\includegraphics[width = 3in,height=2.2in]{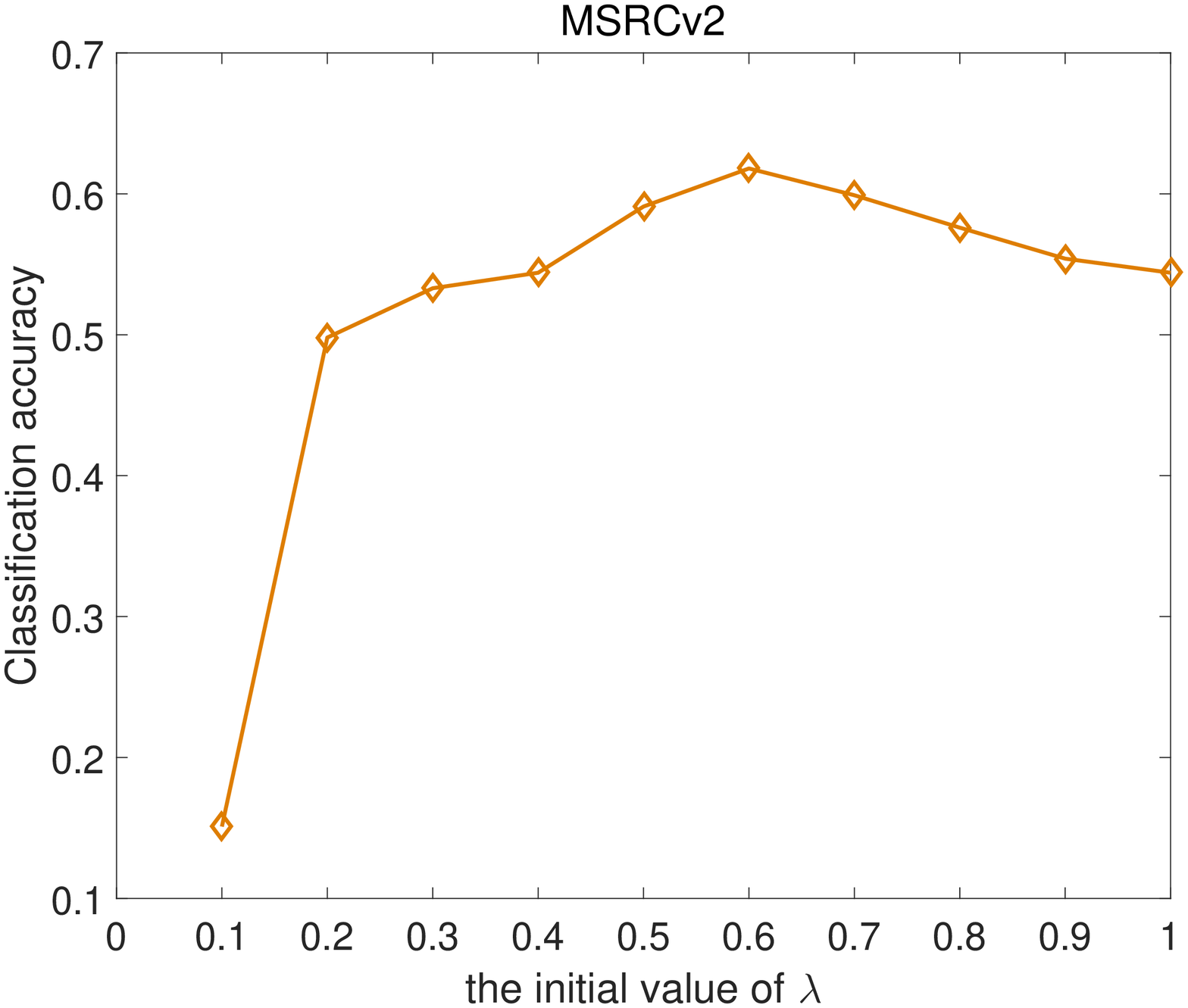}\\
\end{tabular}
\vspace{3mm}
\caption{The classification accuracy of the proposed methods on \emph{Lost} and \emph{MSRCv2} data sets with $C_{max}$ fixed ($C_{max} = 0.01$ on \emph{glass} data set and $C_{max} = 100$ on \emph{MSRCv2} data set respectively)}
\label{Figure5}
\vspace{0mm}
\end{figure*}


\begin{figure*}[!ht]
\centering
\begin{tabular}{cc}
\includegraphics[width = 3in,height=2.2in]{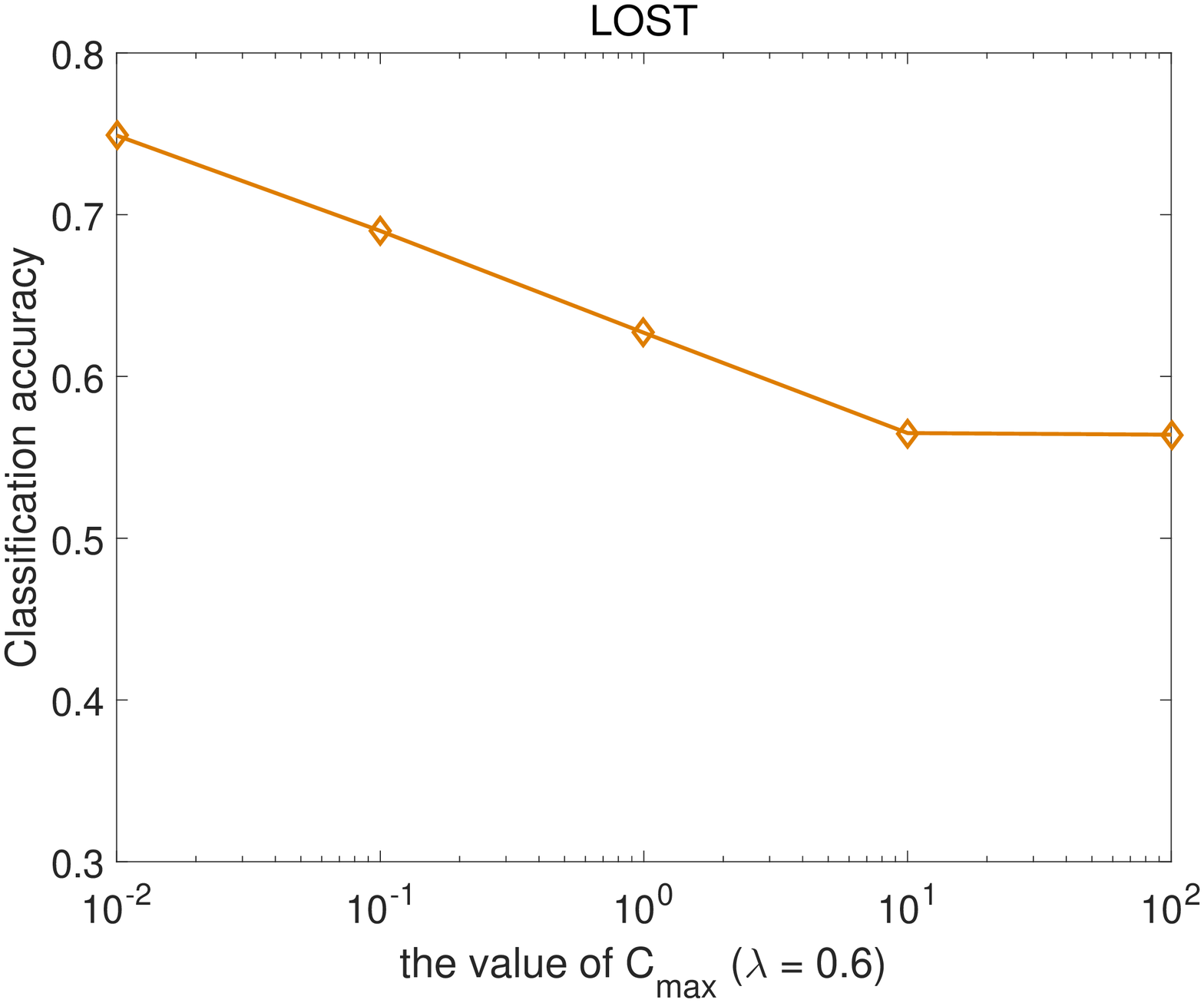}&\includegraphics[width = 3in,height=2.2in]{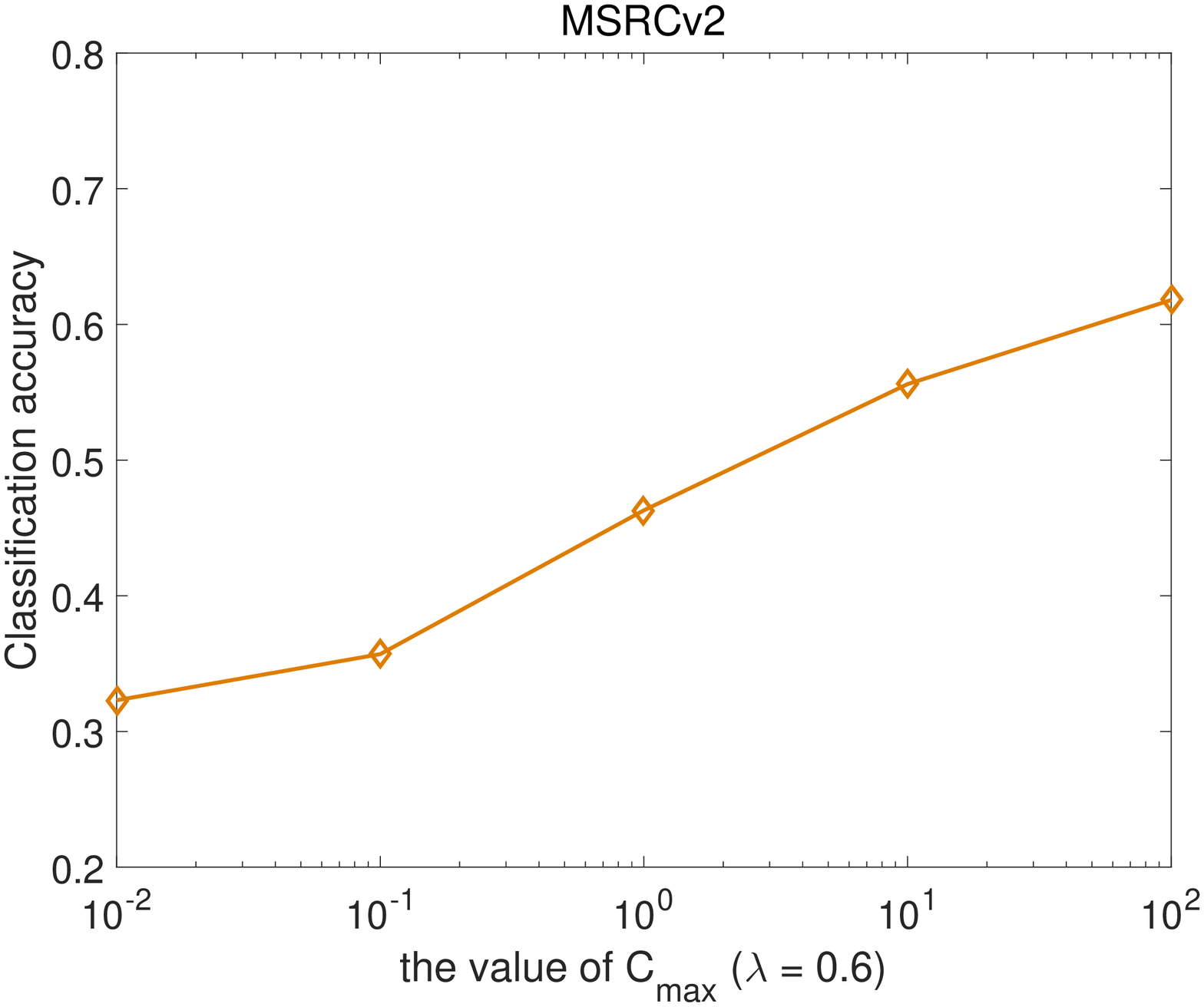}\\
\end{tabular}
\vspace{3mm}
\caption{The classification accuracy of SP-PLL on \emph{Lost} and \emph{MSRCv2} data sets with $\lambda$ fixed ($\lambda = 0.6$)}
\label{Figure6}
\vspace{0mm}
\end{figure*}

The two series of experiments mentioned above powerfully demonstrate the effectiveness of SP-PLL, and we attribute the success to the \emph{easy to hard} self-paced scheme. To learn the instances assigned with high-confidence label firstly can make the reliable label information contribute more to the model. Specifically, during each optimization iteration, we first optimize the assigned labels \textbf{y} and then optimize the classifier parameters $\bm{\Theta}$. When SP-PLL finishes learning from the instances assigned with high-confidence labels in the previous iterations, the unreliable labels associated with the untrained instances have been optimized to become relatively reliable. Thus, the SP scheme can make most data become more reliable before the learning process, which eliminate the noise and increase the reliability of training data to a certain extent. As expected, the experimental results demonstrate the motivation behind our proposed method.

\subsection{Sensitivity Analysis}
The proposed method learns from the PLL examples by utilizing two parameters, i.e. $C_{max}$ (regularization parameter) and $\lambda$ (the self-growth variable). Figure \ref{Figure5} and Figure \ref{Figure6} respectively illustrates how SP-PLL performs under different $C_{max}$ and $\lambda$ configurations. We study the sensitivity analysis of SP-PLL in the following subsection.

\subsubsection{\textbf{Maximum-Margin regularization parameter $C_{max}$}}
The proposed method is based on the maximum-margin strategy, where $C_{max}$ is the regularization parameter to measure the influence of each sample loss on the learning model. Since the $C_{max}$ is usually sensitive to the learning model \cite{Fan:liblinear-JMLR2008}, we empirically set the optimal value of $C_{max}$ on each data set among $\{0.01,\ldots,100\}$ via cross-validation, which is shown in Table \ref{table4}.

\begin{table}[!ht]
\centering
\setlength{\abovecaptionskip}{0pt}%
\setlength{\belowcaptionskip}{5pt}%
\caption{\textbf{The optimal $C_{max}$ for SP-PLL}}
\label{table4}
\begin{tabular}{c|cccc}
\hline
The value of $C_{max}$  & Data Sets           \\ \hline
0.01            & {\emph{Lost}, \emph{FG-NET}, \emph{SoccerPlayer}, \emph{BirdSong}} \\
0.1             & {\emph{Segment}}      \\
10              & {\emph{Glass}, \emph{Letter}}  \\
100             &{\emph{Vehicle}, \emph{MSRCV2}}     \\
\hline
\end{tabular}
\vspace{-1mm}
\end{table}

\subsubsection{\textbf{Self-Growth variable $\lambda$}}
As mentioned above, the main contribution of our method is incorporating the Self-Paced Learning (SPL) scheme into Partial-Label Learning (PLL). The SP parameter $\lambda$ plays an important role in controlling the learning process from easy to hard. The larger the initial value of $\lambda$ is, the more training instances can be learned during the first iterative optimization.
As described in Figure \ref{Figure5}, SP-PLL with a larger value of $\lambda$ tend to achieve poor performance, and we attribute such phenomena to that incorporating more training instances during the first iteration would bring much noise into the learning process, which has negative effect on the final model. Meanwhile, SP-PLL with a smaller value of $\lambda$, which leads to overfitting and weaken the generalization ability of the learning model, also show poor performance. Thus, we empirically guarantee that half of the training instances can be learned in the first iterative optimization. According to Figure \ref{Figure5}, the proposed method achieves desirable performance when the SP parameter $\lambda$ is set to 0.6.

\section{Conclusion}
In this paper, we have proposed a novel self-paced partial-label learning method SP-PLL. To the best of our knowledge, it is the first time to deal with PLL problem by integrating the SPL technique into PLL framework. By simulating the human cognitive process which learns both instances and labels from easy to hard, the proposed SP-PLL algorithm can effectively alleviate the noise produced by the false assignments in PLL setting. Extensive experiments have demonstrated the effectiveness of our proposed method. In the future, we will integrate the SPL technique into PLL framework in a more sophisticated manner to improve the effectiveness and robustness of the model.

\bibliographystyle{IEEEtran}
\bibliography{sp2017}
\end{document}